\documentclass[preprint,12pt]{elsarticle}
\usepackage[T1]{fontenc}
\usepackage{amssymb}
\usepackage{graphicx,times,amsmath}
\usepackage{subfigure}
\usepackage{algorithm}
\usepackage{algorithmic}
\usepackage{subfigure}
\usepackage[utf8]{inputenc}
\usepackage{textcomp}
\usepackage{eurosym}
\usepackage{color}
\usepackage{hyperref}
\usepackage[bottom]{footmisc}
\usepackage{dsfont}
\usepackage{multirow}
\usepackage{booktabs}

\newtheorem{theorem}{Theorem}[section]
\newtheorem{definition}[theorem]{Definition}
\journal{}
\begin{document}
\begin{frontmatter}
\title{Learnability and Robustness of Shallow Neural Networks Learned With a Performance-Driven BP and a Variant PSO For Edge Decision-Making}
\author[1]{Hongmei He}
\ead{h.he@cranfield.ac.uk}
\author[3]{Mengyuan Chen}
\ead{mychen@ahpu.edu.cn}
\author[2,3]{Gang Xu}
\ead{xugang@ahpu.edu.cn}
\author[3]{Zhilong Zhu}
\ead{zhuzhilong919@ahpu.edu.cn}
\author[4]{Zhenhuan Zhu}
\ead{spl.z.zhu@outlook.com}
\address[1]{School of Aerospace, Transport and Manufacturing, Cranfield University, Cranfield, UK}
\address[2] {Key Laboratory of Advanced Perception and Intelligent Control of High-edn Equipement, Ministry of Education, Wuhu, China}
\address[3]{School of Electronic Engineering, Anhui Polytechnic University, Wuhu, China}
\address[4]{Smart Perception Ltd, Milton Keynes, UK}

\begin{abstract}
With the advances in high-performance computing and graphics processing units, deep neural networks attract more attention from researchers, especially in the areas of the image processing and natural language processing. However, in many cases, the computing resources are limited, especially in the edge devices of IoT enabled systems. It may not be easy to implement complex AI models in edge devices. The Universal Approximation Theorem states that a shallow neural network (SNN) can represent any nonlinear function. However, how fat is an SNN enough to solve a nonlinear decision-making problem in edge devices? In this paper, we focus on the learnability and robustness of SNNs, obtained by a greedy tight force heuristic algorithm (performance driven BP) and a loose force meta-heuristic algorithm (a variant of PSO). Two groups of experiments are conducted to examine the learnability and the robustness of SNNs with Sigmoid activation, learned/optimised by KPI-PDBPs and KPI-VPSOs, where, KPIs (key performance indicators: error (ERR), accuracy (ACC) and $F_1$ score) are the objectives, driving the searching process. An incremental approach is applied to examine the impact of hidden neuron numbers on the performance of SNNs, learned/optimised by KPI-PDBPs and KPI-VPSOs. From the engineering prospective, all sensors are well justified for a specific task. Hence, all sensor readings should be strongly correlated to the target. Therefore, the structure of an SNN should depend on the dimensions of a problem space. The experimental results show that the number of hidden neurons up to the dimension number of a problem space is enough; the learnability of SNNs, produced by KPI-PDBP, is better than that of SNNs, optimized by KPI-VPSO, regarding the performance and learning time on the training data sets; the robustness of SNNs learned by KPI-PDBPs and KPI-VPSOs depends on the data sets; and comparing with other classic machine learning models, ACC-PDBPs win for almost all tested data sets.
\end{abstract}

\begin{keyword}
Universal Approximation Theorem \sep Performance-Driven BP learning \sep Incremental Approach \sep Variant PSO \sep Learnability and Robustness of Shallow Neural Networks \sep Edge Decision Making.
\end{keyword}
\end{frontmatter}
\section{Introduction}\label{sec:Introduction}
An artificial neuron network (ANN) is a computational model, mimicking the structure and functions of biological neural networks or human brains. It usually consists of an input layer, some hidden layers, and an output layer. A shallow neural network (SNN) is the neural network with only one hidden layer, containing a finite number of neurons. An ANN provides a straightforward approach to creating the relations between input attributes and the output based on a limited set of data, instead of an exact mathematical function that we may not be able to create. The ability to learn by samples makes ANNs very flexible and powerful. Although there exists bias to the real relation between inputs and outputs, ANN is still an excellent approach to solving many nonlinear mapping problems.

Deep neural networks (DNNs) have been successfully applied in two main areas: image processing and speech recognition. Especially, deep convolutional neural networks (DCNNs) have brought about breakthroughs in videos \cite{KarpathyTS2014,Wang2018}, image processing \cite{ChenPK2017}, object detection \cite{RenHG2017},  as well as audio \cite{LeeLP2009} and speech recognition \cite{ZhangW2017}. The properties of composition hierarchies of images, speech, and text promote the capacities of deep neural networks. However, we cannot always see the semantics of higher-level features in many real-world cases as in image and acoustic modeling. A DCNN comprises multiple layers of feature representations and a fully connected neural network (i.e., an SNN) in the last two layers. While the convolutionary and pool layers are essential to represent the features of inputs, the final two layers of a fully connected neural network are important to the performance of the DCNN.

Generally, ANN training is a supervised learning process. It is to adjust the network's state in response to data. The learner (learning or training algorithm) receives a set of training samples, consisting of ordered pairs of the form ($\vec{x}$,$y$) (called labelled samples), where $\vec{x}$ is an input vector to the neural network ($\vec{x} \in X$) and $y$ is an output ($y \in Y$). The backpropagation (BP) algorithm is a classic training algorithm of ANNs. Blum and Rivest \cite{Blum1992} proved that training a 2-layer, 3 nodes and $n$ inputs neural network with the BP algorithm is NP-Complete. The significant barrier to blocking the applications of deep neural networks is the computing complexity, although it shows a great attraction in solving complex nonlinear problems. With the strong computing capability of GPU, deep learning for 2-20 depth networks is successful (e.g., Google AlphaGo). The success of deep learning in image and acoustic modeling benefits from GPU computing. However, in many cases, we may not need to use a GPU, or even we may not have a GPU to support the complex calculation. For example, in an IoT enabled system, the computing capacity and memory size are limited in edge devices. More importantly, when edge intelligence needs to be implemented with adaptivity, online learning is required to have real-time performance.

The Universal Approximation Theorem (UAT), first with Sigmoid activation function, proved by Cybenkot in 1989 \cite{Cybenkot1989}, states that an SNN with a non-polynomial activation function can approximate any function, i.e., can in principle learn anything \cite{Hornik1991,Leshno1993}. UAT indicates that a shallow neural network (SNN) could be enough to solve any nonlinear approximation problems. However, there is a little research focusing on the algorithm learnability and robustness for UAT. Given a set of data, how many hidden neurons in an SNN are enough to achieve high performance for decision making? Especially in edge computing, computing complexity is a critical challenge. It may not be easy to implement complex AI models in edge devices. We expect to use a small and simple model to well solve complex problems with a minimum cost. Comparing to deep learning, SNNs may not lose the ground truth of the relationship between inputs and outputs, represented by the training data, in terms of UAT.

In principle, the non-linearity of the training data decides the number of hidden neurons in an SNN. If the relationship between inputs and outputs is strongly nonlinear, then the number of hidden neurons could be larger. From an engineering perspective, sensors in a monitoring system are always justified for necessity. In other words, all input attributes from sensors are essential for a targeted problem. Therefore, feature selection is not considered in this work. However, the non-linearity of data may be strong.

Improving learning performance for all ANN applications is essential. Usually, there are four kinds of approaches to improving the performance of ANNs: (1) improving the quality of data. The output of a learning algorithm is a function $h\in H$, which maps the relationship between the inputs of all training samples to the labeled outputs $y$. Therefore the quality of training samples is vital for training a neural network; (2) improving learning algorithm, for which many notable algorithms have been developed in addition to the classic backpropagation (BP) algorithm; (3) neural network tuning, for which some evolutionary algorithms were developed to optimize the parameters and neural network structures; (4) using ensembles to improve the robustness of decision making or classification.

The BP algorithm uses a greedy tight force (gradient descent) to drive a network from one state to another, but it may fall in a spurious local optima. A Particle Swarm Optimization (PSO) is a meta-heuristic global optimization paradigm, using a loose force to drive the search in the problem space. It has gained prominence in the last two decades due to its ease of applications in complex multidimensional problems \cite{Sengupta2019}.

In this paper, we aim to examine the learnability and the robustness of SNNs with Sigmoid activation, through improving the learning algorithm and neural network tuning, given a data set. Hence, a performance-driven backpropagation (PDBP) algorithm and a variant of particle swarm optimization (VPSO) are developed to learn or optimize the weights of an SNN for nonlinear decision making, respectively. Our previous research \cite{He2018} observed the impact of hidden neuron numbers on the performance of SNNs for different data sets, using an incremental approach to adding a hidden neuron into an SNN gradually. It was shown that when the number of hidden neurons in an SNN for general data sets reaches about half or more of the input number in the problem space, the performance is not improved further or has a tiny change. For a better linear data set, the number of needed hidden neurons could be smaller. For example, for the Wisconsin Breast Cancer (WBC) data set, the performance does not improve much, when the number of hidden neurons gets larger than 2, and even when an additional hidden layer is added into the SNN. Therefore, in the experiments, the impact of hidden neuron number in an SNN is observed from one to the input number of the problem space.

To examine the learnability and robustness of an SNN,  we conduct two groups of experiments with PDBP and VPSO on the two benchmark databases, WBC \cite{Wolberg1990} and SMS spams \cite{Almeida2011}, from the UCI machine learning repository \cite{Dua2017}. The performance of SNNs obtained by PDBP and VPSO on more data sets from the UCI machine learning repository will be compared with other machine learning models in the literature.

The rest paper is organized as follows: Section 2 will survey existing work on the algorithm learnability in theory and practice; Section 3 proposes the PDBP algorithm; Section 4 introduces the VPSO algorithm; Section 5 conducts the two groups of experiments to examine the learnability and robustness of SNNs with PDBP and VPSO; and finally, conclusions are presented in Section 6.

\section{Existing work}
The learnability of neural networks has been addressed from a theoretical perspective and practice in the last two decades. A formal definition of 'learning' was given in \cite{Anthony1999}:

\begin{definition}
Suppose that $H$ is a class of functions that map from a set $X$ to $[0,1]$. A learning algorithm $L$ for $H$ is a function, mapping from the set of all training samples $z_N$ to $H$:
\begin{align}
L : \bigcup_{N=1}^{\infty}Z_N \rightarrow H.
\end{align}
$L$ is to produce a $h\in H$, so that $E_P(h)$ < $E_{opt_P}(H)$ + $\epsilon$, where, $h$ is the output of the learner, $E_P(h)$ is the approximated error produced by the hypothesis, $E_{opt_P}(H)$ is the error produced by the optimal hypothesis in $H$, and $\epsilon \rightarrow 0$.
\end{definition}
Namely, the aim of learning is to produce a function $h$ in $H$ that has near minimal error $E_P(h)$. $H$ is learnable if there is a learning algorithm for $h\in H$, which has the approximated minimal error $E_P(h)$.


Anthony and Bartlett \cite{Anthony1999} approved the learnability of a neural network, representing finite function classes. A learning algorithm $L_z$ chooses the hypothesis function with minimal error on training samples $z$, as shown in Eq. \ref{eq:minerr}:
\begin{align}\label{eq:minerr}
\hat{E}_z(L_z)=min_{h\in H}\hat{E}_z(h)
\end{align}
An algorithm $L_z$ is an efficient consistent-hypothesis finder for the graded binary class $H$ = $\bigcup h_{n}$ if, given $N$ training samples $z$ for a target function in $H$, $L_z$ halts in time polynomial of $N$ and $n$, and returns $h = L(z) \in H$ such that $\hat{E}_z(h)$ = 0.

Sharma et al. \cite{Sharma2018} studied the learnability of learned neural networks under various settings of neural networks, and provided the definition of learnability of neural networks. Assume a multi-class classification problem with $C$ classes and let $D$ denote a distribution over the inputs $\vec{x} \in \mathbf{R}^d$. Assume an ANN was obtained with $N$ independent samples $\vec{x}_i\in D_{tr}$, $i=1..N$, $D_{tr}$ is the training set, corresponding estimates obtained by the learned ANN model $\hat{\mathcal{M}}$ are $\hat{\mathcal{M}}(\vec{x}_1)$, ..., $\hat{\mathcal{M}}(\vec{x}_n)$, then the learnability ($\mathcal{L}(\mathcal{M})$) of an ANN learned with the algorithm is defined as:
\begin{align}
\mathcal{L}(\mathcal{M})= \frac{1}{N}\sum_{\vec{x}_i\in D_{tr},i=1}^N
\mathbf{1}_{\mathcal{M}(\vec{x}_i)=\hat{\mathcal{M}}(\vec{x}_i)}\times 100\%.
\end{align}
Namely, the learnability of a classifier, trained by a learning algorithm, is the accuracy of $\hat{\mathcal{M}}$ on the training data set $D_{tr}$, and $\mathcal{L}(\mathcal{M})$ implicitly depends on $D_{tr}$, the architecture of the ANN, and the learning algorithm, used to learn the ANN model $\mathcal{M}$, as well as sample number $N$. However, they did not consider the learning time.
Zhong et al. \cite{Zhong2017} investigated recovery guarantees for SNNs with both sample complexity and computational complexity linear in the input dimension and logarithmic in the precision, and their study showed that tensor initialization followed by gradient descent is guaranteed to recover the ground truth with a certain sample complexity and computational complexity for smooth homogeneous activation functions with a high probability. Kim et al. \cite{Kim2019} investigated the concatenation of additional information as supplementary axes to improve the learnability of neural networks. To measure the relative importance of each sample in a training set,  Lee et al. \cite{Lee2019} raised the concept of sample-wise learnability within a deep learning context, and proposed a measure of the learnability of a sample $(\vec{x}_c, y_c)$ with a given deep neural network (DNN) model:
\begin{align}
L_f(\vec{x}_c, y_c) = \mathbf{E}[\frac{1}{T}\Sigma_{t=1}^{T} f_{y_c}^{(t)}(\vec{x}_c)],
\end{align}
where $T$ denotes the total number of training steps, $f_{y_c}^{(t)}(\vec{x}_c)$ is the probability that the model predicts the label of $\vec{x}_c$ as $y_c$. If $\vec{x}_c$ is easily learnable, the value of $f_{y_c}^{(t)}(\vec{x}_c)$ increases rapidly to 1 as the training step $t$ increases; otherwise, the value of the probability $f_{y_c}^{(t)}(\vec{x}_c)$ remains small and so does the value of $L_f(\vec{x}_c, y_c)$.
Obviously, the learnability of an ANN is related to the prediction accuracy and training time. However, it should be related to the final prediction performance and the total learning time, rather than the performance during the learning process. After all, we expect a learned ANN with a high prediction accuracy and a short training time, no matter what the learning process is.

Ge et al. \cite{Ge2018} analyzed the population risk of the standard squared loss and designed a non-convex objective function ($G$) for learning an SNN. Their experimental results show that a stochastic gradient descent on $G$ provably converges to the global minimum and determines the ground-truth parameters. Li and Yuan \cite{LY2017} provided a formal analysis of the convergence of stochastic gradient descent (SGD) based techniques for two-layer feed-forward networks with the ReLU activation. Song et al. \cite{Song2017} provided a comprehensive lower bound, which implies that any statistical query algorithm, including all known variants of stochastic gradient descent algorithms with any loss function, to learn an SNN for a wide class of activation functions and inputs drawn from any log-concave distribution, needs an exponential number of queries even using tolerance inversely proportional to the input dimensions.

Bengio et al. \cite{Bengio2009} proposed a curriculum learning, in which a model is learned gradually in the order of increasing entropy of training samples. A curriculum determines a sequence of training samples, which substantially corresponds to a list of samples, ranked in ascending order of learning difficulty. Namely, the order of the samples, fed to an ANN model, will affect the curriculum learning. The main challenge in using the curriculum learning strategy is that it requires the identification of easy and hard samples in a given training data set. To overcome the drawback of curriculum learning, Kumar et al. \cite{Kumar2010} proposed a self-paced learning system, where the curriculum is determined by the learner's abilities rather than being fixed by a teacher. Combining curriculum learning and self-paced learning, Jiang et al. \cite{Jiang2015} proposed a Self-Paced Curriculum Learning.

Zhang et al. \cite{Zhang2017} investigated the learnability of fully connected neural networks. They proposed a boostNet algorithm by using the AdaBoost approach \cite{Freund1997} to construct the network. The basic idea of the algorithm is that the algorithm trains a shallower network (e.g., an $m$-1 layer network) with an error rate slightly better than random guessing, then adds it to the classifier to construct an m-layer network. The experimental results have shown that the prediction error obtained by the proposed algorithm is less than that obtained by the BP algorithm.

Arora et al. \cite{Arora2014} proposed an algorithm to learn a generative deep network model, and most networks were obtained in polynomial running time. The algorithm uses layerwise learning, based upon a novel idea of observing correlations among features and using these to infer the underlying edge structure via a global graph recovery procedure.

Janzamin et al. \cite{Janzamin2016} proposed an algorithm based on tensor decomposition for guaranteed training of two-layer neural networks under various settings and provided risk bounds for the proposed method with a polynomial sample complexity in the relevant parameters, such as input dimension and number of neurons.

The training of neural networks can be a non-convex optimization problem. There was much research on the weight optimization of neural networks. Kajornrit \cite{Kajornrit2015} examined the learning performance of neural networks with meta-heuristic algorithms, including Simulated Annealing, Direct search, and Genetic Algorithm. They executed the BP algorithm on the neural networks obtained by these meta-heuristic algorithms on the tested databases, and GA achieved the best performed neural network, compared with other meta-heuristic algorithms. Nawi et al. \cite{Nawi2015} proposed a Cuckoo Search algorithm, inspired by Cuckoo bird's behavior, to train the Elman recurrent network and the backpropagation Elman recurrent network with fast convergence.

David and Greental \cite{David2014} used a GA to optimize Deep Neural Networks, but they did not implement it. As we argued in the introduction, the success of deep neural networks benefited from GPU. Assuming the complexity of deep neural networks is $\kappa$, the number of evolutionary iterations is $\mathcal{I}$, and the population size $P$, and we have $u$ processors to parallelise the GA, the complexity of the optimisation process of deep neural network is $O(\mathcal{I}\kappa)$. Hence, the complexity of GA on each processor is $O(\mathcal{I}\kappa P/u)$, which is intolerant in an edge device. Zhang et al. \cite{Zhang2018} proposed a dynamic neighborhood learning-based gravitational search algorithm (GSA). This approach can improve search performance in convergence and the diversity of evolutionary optimization. However, our preliminary experiments show that GSA cannot converge very well for a high-dimension optimization problem.

There was also some research on incremental approaches. For example, Bu et al. \cite{Bu2015} proposed an incremental backpropagation model for training neural networks by adapting the parameters and the neural structure and used the Singular Value Decomposition on the weight matrix to reduce some redundant links. The final neural network is not a fully connected ANN.  He et al. \cite{He2016} used the incremental approach in the order of decreasing information gains of all features to select features for a Support Vector Machine based on the RBF kernel for spam detection.
\section{Performance Driven BP for SNN training}\label{sec:BP}
Neural network learning is to find the optimal weights so that the network approximates the function of representing the given data as closely as possible. Namely, given a training set with $N$ samples, ${(\vec{x}_1,y_1),...,(\vec{x}_N,y_N)}$, it is to minimize the error function of the network, defined as \cite{Rojas1996}
\begin{align}\label{eq:error}
\mathcal{E} =\frac{1}{2}\sum_{i=1}^{N}{\|o_i-y_i\|^2},
\end{align}
where, $o_i$ is the output of the SNN for input sample $\vec{x}_i$, $y_i$ is the target output.
The basic idea of the BP algorithm is to use error backpropagation to update weights in a fixed structure of ANN. All weights, including biases, are initialized with a small real value in [-0.5,0.5], the output of the ANN is calculated through the feed-forward process, the average error $\varepsilon$ is calculated with Eq. (\ref{eq:error}) divided by $N$, then the error is back-propagated from the output layer layer by layer, and the weights are updated with the gradient of the error. The process is repeated until the criterion is met.

In the general BP algorithm, the stop criterion depends on the average error $\varepsilon$. The question is how small the average error is sufficient. If a too small threshold of $\varepsilon$ is set, then the number of learning iterations could be very large. Hence, usually, a maximum number of iterations is set, in case the average error cannot converge to the specified value. A small error criterion could increase the complexity of training. Moreover, a neural network system may have an overfitting problem. To avoid overfitting, usually, a small data set could be used to validate the performance during the training process. Once the error produced by the SNN on the validation data is increasing, while the error produced by the SNN on training data is still decreasing, the training process will be stopped. This also increases the computing complexity of the training process.

The goal of neural network training is to gain an SNN that is a highly performed decision maker. The initial experimental results show that when average error arrives at a certain value, the performance (e.g., accuracy) could not be improved further. Therefore, one of the stop criteria can be set to evaluate the performance. The key performance indicators (KPIs) include Accuracy (ACC), $F_1$-score, Area under the curve of ROC (AUC), True Positive Rate (TPR), and True Negative Rate (TNR), etc. Namely, the learning process will not be stopped until a specific KPI of the learned network has not been improved for a certain number ($\tau$) of iterations (called convergence tolerance), as shown in Algorithm \ref{alg:PDBP}. When the average error is used as the performance measure, the line 10 in Algorithm \ref{alg:PDBP} should be $perf < best_p$, and the BP learning process becomes a minimising process. In Algorithm \ref{alg:PDBP}, $D(X,Y)$ is the training data with pairs of input samples $X$ and corresponding labels $Y$, the function of calPerformance($Y$, $\tilde{Y}$) is to calculate the specified performance indicator that is used to drive the convergence process.
\begin{algorithm}[ht]
\caption{PDBP(D(X,Y),Net, $\tau$)}\label{alg:PDBP}
\begin{algorithmic}[1]
\STATE Initialise(Net);
\STATE $t$ = 0, $k$ = 0;
\STATE $perf$ = 0;
\WHILE{(t < MAX\_IT)}
    \STATE $\tilde{Y}$=feedforward($D_x$);
    \STATE backpropagation($Y$, $\tilde{Y}$);
    \STATE $Net$ = updateWeight($Net$);
    \STATE $best\_p = perf$;
    \STATE $perf$ = calPerformance($Y$, $\tilde{Y}$);
    \IF  {$(perf>best\_p)$}
        \STATE $k$ = 0;
        \STATE $best\_p=perf$;
    \ELSE
        \STATE $k$ = $k$+1;
        \IF {($k$>$\tau$)}
            \STATE break;
        \ENDIF
    \ENDIF
    \STATE $t = t$+1;
\ENDWHILE
\end{algorithmic}
\end{algorithm}
Assume there are $m$ hidden neurons in an SNN. For a data set with $n$ input attributes, all $n$ input attributes are fully linked to all the $m$ hidden neurons, respectively, and the $m$ hidden neurons are fully connected to the output neuron. Each neuron has a bias $b$.  For an SNN to solve a decision making problem, the total number of weights can be calculated as Eq. (\ref{eq:dimensions}):
\begin{align}\label{eq:dimensions}
d = m(n+1)+m+1 = m(n+2)+1.
\end{align}
For each weight, BP will calculate the feedback through computing the error, of which the computing complexity is $O(N)$ for $N$ samples. Therefore, for each iteration, the computing complexity is $O(dN)$ for $d$ weights.
\section{A VPSO for the Optimisation of an SNN}
The canonical particle swarm optimizer is based on the flocking behavior and social co-operation of birds and fish schools and draws heavily from the evolutionary behavior of these organisms \cite{Sengupta2019}. A variant of particle swarm optimiser (VPSO) is developed to search the optimal weights in an SNN. The pseudo-code is shown in Algorithm \ref{alg:VPSO}. A population of particles are initialized, and each particle represents an SNN. In a evolutionary loop, the outputs of SNNs are calculated, then the local best and global best are found and saved, and they are used to update the velocity and the value of each particle. The topology of all particles is changed in the period of $T$. If the global best performance is not changed for a certain number ($\tau$) of iterations or the generation reaches to the maximal number of iterations, the process will be stopped.
\begin{algorithm}[htp]
\caption{VPSO(D(X,Y),$m$)}\label{alg:VPSO}
\begin{algorithmic}[1]
\small
\STATE $t$ = 0, $k$ = 0;
\STATE $\{N_{pop}, \Gamma, T, C_1,C_2, \chi, \theta\}$=initParameters();
\STATE $\{\mathcal{P}_{1..N_{pop}}, \mathcal{V}_{1..N_{pop}}\}$=InitParticles($d$,$N_{pop}$);
\WHILE{($t$<max\_iter)}
    \STATE $y_{1..\mathcal{N}}$=getNNouts($\mathcal{P}_i$, $X$);
    \STATE $perf_{1..\mathcal{N}}$ = calPerform($\theta_y$, $y_{1..\mathcal{N}}$, $Y_{1..\mathcal{N}}$);
	\STATE $left_{1..N_{pop}}$  = get\_left();  \%left neighbor of each position
	\STATE $right_{1..N_{pop}}$ = get\_right(); \%right neighbor of each position
	\STATE $\{sBestValue_{1..N_{pop}}(t),sBestPosition_{1..N_{pop}}(t)\}$ = get\_sBest(); \%Eq.(\ref{eq:sbest})
    \STATE $\{pBestValue_{1..N_{pop}}(t),pBestPosition_{1..N_{pop}}(t)\}$ = get\_pBest(); \%Eq.(\ref{eq:gbest})
    \IF {($gBestValue$ unchanged)}
        \STATE $\tau = \tau+1$;
        \IF {($\tau$ = $\Gamma$)}
            \STATE $lastStep$ = $t$, break;
        \ENDIF
    \ENDIF
    \STATE $t = t$+1;
    \IF {($t$ mod $T$) == 0)}
        \STATE reorder($\mathcal{P}_{1..\mathcal{N}}(t)$);
    \ENDIF
    \STATE $M$ = get\_mass();
    \STATE $\varpi$(t) = get\_inertiaweight($M$);\%Eq.(\ref{eq:inertiaweight});
	\FOR {(i=0... $N_{pop}$)}
	    \FOR {( $j$=0; $j$< $d$; $j$++)}
    	  \STATE $r_1$ = rand()$\in(0,1]$; $r_2$ = rand()$\in(0,1]$;
          \STATE $V_{ij}(t+1)$ = update\_velocity($r_1,r_2,V_{ij}(t)$);\%[-1,1], Eq.(\ref{eq:variant-v})   	
        \ENDFOR
    \ENDFOR
	\FOR {(i=0... $N_{pop}$)}
        \FOR {(j = 0 ... $d$)}
    		\STATE $\mathcal{P}_{ij}$ = $\mathcal{P}_{ij} + \mathcal{V}_{ij}$;
    			\IF {($(\mathcal{P}_{ij} > 1$ or $(\mathcal{P}_{ij} < -1)$)}
    			    \STATE $\mathcal{P}$ = rand()$\in[-1,1]$;
    			    \STATE $\mathcal{V}_{ij}=0.1\mathcal{P}_{ij}$;
                \ENDIF
        \ENDFOR
    \ENDFOR
\ENDWHILE
\end{algorithmic}
\end{algorithm}
\subsection{Positions of particles}
VPSO employs a swarm of particles to search their optimal position in a multidimensional search space. Each particle represents a potential solution (i.e., the weights of an SNN) and evolves in terms of the experiences of its neighbors and itself. The number $d$ of dimensions of a particle can be calculated with Eq. (\ref{eq:dimensions}). A particle is an array of real numbers in [0,1], $\mathcal{P}$ = $\{p_1,...,p_d\}$, $-1\leq p_i\leq 1$, and it represents Matrix (\ref{w-matrix}).
\begin{equation}
\begin{bmatrix}\label{w-matrix}
w_{11}, &w_{12}, &..., &w_{1n}, &b_1\\
w_{21}, &w_{22}, &..., &w_{2n}, &b_2\\
     ...& ...  , &..., &...,    &\\
w_{m1}, &w_{m2}, &..., &w_{mn}, &b_m\\
w_{y1}, &w_{y2}, &..., &w_{ym}, &b_y\\
\end{bmatrix}
\end{equation}
For example, assume there are eight input attributes in a training data set and four hidden neurons in an SNN, the dimension number $d$ of a particle is  41 (= $4\times (8+2)+1$). Obviously, VPSO is used to solve a high dimensional optimization problem.
\subsection{Update Equations of the Standard PSO}
In the standard PSO, the velocity is updated with Eq. (\ref{eq:velocity}):
\begin{align}\label{eq:velocity}
v_{ij}(t+1) = \varpi  v_{ij}(t) + & C_1 r_1(t) (pBest_{ij}(t)- p_{ij}(t))\nonumber\\
                                + & C_2 r_2(t) (gBest(t) - p_{ij}(t)),
\end{align}
where, $v_{ij}(t)$ is the velocity of the $i$-th particle in the $j$-th dimension at iteration t, $\varpi$ is the inertia weight, $r_1$ and $r_2$ are independent and identically distributed random numbers, $C_1$ and $C_2$ are the cognition and social acceleration coefficients, $pBest_{ij}(t)$ and $gBest(t)$ represent the personal and global best positions. Then, the corresponding particle is updated with Eq. (\ref{eq:position}):
\begin{align}\label{eq:position}
p_{ij}(t+1) = p_{ij}(t) + v_{ij}(t+1).
\end{align}
\subsection{Inertia Weight}
The inertia weight $\varpi$ is a control parameter for the swarm velocity. It can be a constant, linear time-varying, or even nonlinear temporal dependencies \cite{Shi1998b,Naka2001}. The value of the inertia weight could affect the convergent behavior and the balance between exploitation and exploration.  $\varpi \geq 1$ implies the swarm velocity increases over time towards the maximum velocity $\mathcal{V}_{max}$. The inertia weight can be implemented through decreasing the value of $\varpi$ from a preset high value of $\varpi_{max}$ to a low of $\varpi_{min}$ \cite{Suganthan1999,Ratnaweera2003}. Conventionally, $\varpi_{max}=0.9$ and $\varpi_{min} = 0.4$.
An Adaptive Inertia Weight is proposed, borrowing the mass concept in Gravitational Search Algorithm \cite{Zhang2018}:
\begin{align}\label{eq:mass}
mass_i(t)=\frac{f_i(t)-worst(t)+\epsilon}{best(t)-worst(t)+\epsilon},
\end{align}
where, $f_i(t)$ represents the fitness value of particle $\mathcal{P}_i$ at time $t$, and  $\epsilon$ is a very small value (e.g. 0.001). This makes the equation available for calculation and $mass(t)$ tend to 1, when all particles converge to one fitness, namely, $f_i(t)$ = $worst(t)$ = $best(t)$. Hence, $0<mass(t)\leq 1$. For a maximisation problem, the $best(t)$ and $worst(t)$ are defined as follows:
\begin{align}
best(t) = max(f_1(t),...f_{\mathcal{N}}(t));
worst(t) = min(f_1(t),...,f_{\mathcal{N}}(t));
\end{align}
Hence, the adaptive inertia weight can be calculated with Eq. (\ref{eq:inertiaweight}):
\begin{align}\label{eq:inertiaweight}
\varpi_i(t) = \varpi_{max} - 2.164(\varpi_{max}-\varpi_{min})\frac{e^{mass_i(t)}-1}{e^{mass_i(t)}+1},
\end{align}
where, $\varpi_{max}$ and $\varpi_{min}$ are the conventional values, but the inertia weight is decreasing from 0.9 to 0.4 as the mass value increases from 0 to 1.
\subsection{Cognitive and Social Acceleration Coefficients}
The acceleration coefficients $C_1$ and $C_2$, multiplying with random vectors $r_1$ and $r_2$, respectively, control stochastic changes on the velocity of the swarm.
$C_1$ and $C_2$ can be viewed as the weights, weighting how much a particle should move towards its cognitive attractor ($pBest$) or its social attractor ($gBest$). The personal best value $pBest$ is defined as Eq. (\ref{eq:pbest}):
\begin{align}\label{eq:pbest}
 pBest_i(t) = max(f_i(t),pBest_i(t-1)).
\end{align}
The $gBest$ is defined as Eq. (\ref{eq:gbest}):
\begin{align}\label{eq:gbest}
gBest(t) = max(pBest_1(t),...,pBest_{\mathcal{N}}(t),gBest(t-1)),
\end{align}
where, $\mathcal{N}$ is the population size.

In VPSO, the topology of particles defines the neighborhood of a particle. Each particle holds the best value $sBest$ among its neighbors and itself historically. This is equivalent to a small swarm of learning with neighbors and makes VPSO suitable for high-dimension swarm optimization problems, well mimicking the behavior of flocking birds.

\begin{align}\label{eq:sbest}
gsBest_i(t) = max_{\mathcal{P}_i'\in Neighbors(\mathcal{P}_i)}(f_{i'}(t),f_i(t), sBest_i(t-1)), i=1,.., \mathcal{N}.
\end{align}
The global best ($gsBest$) is defined as Eq. (\ref{eq:gsBest}):
\begin{align}\label{eq:gsBest}
gsBest(t) = max(sBest_1(t),...,sBest_{\mathcal{N}}(t),gsBest(t-1)),
\end{align}
where, $\mathcal{N}$ is the swarm size.

Assume the topology of particles is set to a circle, then the $sBest$ is the best among the left neighbor of $\mathcal{P}_i$, $\mathcal{P}_i$ itself, and the right neighbor of $\mathcal{P}_i$. In a circle, if $i$=0, then $left$=$\mathcal{N}-1$, else, $left$ = $i$-1; $right$ = $(i+1) mod \  \mathcal{N}$.

The topology can be changed periodically. Hence, the neighborhood of a particle is changed. The topology period $T$ can be proportional to the swarm size $\mathcal{N}$.  All particles are reordered in a circle periodically in terms of some strategies.
\subsection{Constriction Factor and Particle Update}
A constriction coefficient $\chi$ can be set to ensure optimal trade-off between exploration and exploitation \cite{Clerc1999,Clerc2002}. The constriction co-efficient was developed from eigenvalue analyses of computational swarm dynamics in \cite{Clerc1999}, as shown in Eq. (\ref{eq:chi}):
\begin{align}\label{eq:chi}
\chi = \frac{2}{2 - \varphi - \sqrt{\varphi (\varphi - 4)}},
\end{align}
where, $\varphi = C_1 + C_2$.  For $C_1$=2.05, $C_2$=2.05, $\chi$ = 0.7298.

The update equation of velocity becomes to Eq. (\ref{eq:variant-v}):
\begin{align}\label{eq:variant-v}
v_{ij}(t+1) = \chi (\varpi(t) v_{ij}(t) + &C_1 r_1 (sBest_{ij}(t) - p_{ij} (t)) \nonumber\\
                                        + &C_2 r_2 (gsBest_{ij}(t) - p_{ij}(t))),
\end{align}
\subsection{Some Parameters and Strategies of VPSO}
The outputs of an SNN for a data set $D_x$  can be expressed as $Y$ = SNN($\mathcal{P}$, $D_x$), where $\mathcal{P}$ = $\{p_1,...,p_d\}$, representing the weights of the SNN (Matrix (\ref{w-matrix})), $D_x={\vec{x}_1,...,\vec{x}_N}$. The estimated outputs should be as close as possible to the desired outputs that have been labeled in the data set. There are many performance indicators to measure an SNN. Hence, the fitness measure can be one of the key performance indicators. Our preliminary experiments show that when TPR is used as the fitness of VPSO, VPSO always goes to the extreme local optimum (i.e., TPR =1, but TNR =0). When AUC is used as a fitness of VPSO, for WBC, the local optimum (TPR=1, TNR=0) always appears, and for SMS, the local optimum (TNR =1, TPR is close to 0) is always reached. However, in both cases, the values of AUC are very high. This is similar to the case in the optimization of a linguistic attribute hierarchy, obtained by the Genetic Algorithm in \cite{HL2014}. A deep study of AUC will be discussed in the future. PDBPs with TPR and AUC as drivers do not have this problem. It is because the evolution of PDBP is essentially based on error, and a KPI is just used to drive the convergence process. Therefore, the three KPIs, such as ERR, ACC, and F1, are used as fitness measures of VPSO.

 The dimension number $d$ of a particle is linearly changed as the number of hidden neurons changes, as shown in Eq. (\ref{eq:dimensions}). The search space is decided by the dimension number $d$ of a particle. Hence, the population size is set to 2$d$, and the maximum iteration number is set to 20,000. To save the evolutionary time, we also set a convergence tolerance ($\tau$). If the global best fitness ($gsBest$) has not been changed for $\tau$ iterations, the VPSO will be stopped. Similar in PDBP, the tolerance number is proportional to the dimension number, and set to $20d$. For diversity, the topology of particles will be randomly rearranged in a circle in each $T$ iterations, $T = \tau/2$.
\section{Experiments and Evaluation}
\subsection{Experiment setup}
\begin{figure}[htp]
\centering
\includegraphics[height=2.1in]{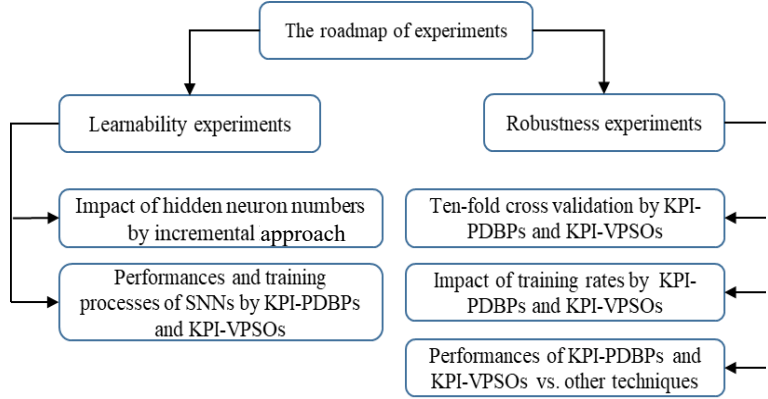}
\caption{The roadmap of experiments} \label{fig:roadmap}
\end{figure}

The test platform is a desktop with Windows 10 and Intel (R) Core (TM)2 Duo CPU T7300 @2GHZ 2GB memory. The algorithms were implemented in the platform of Eclipse C++. The parameters in PDBP and VPSO are set in Table \ref{tab:para}:
\begin{table}[ht]
\centering
\caption{Parameters of PDBP and VPSO}\label{tab:para}
\begin{tabular}{l|c|l}
  \hline
  parameters                            & values        & algorithms    \\
  \hline
  max\_iterations                       & 20,000        & PDBP, VPSO    \\
  Convergence tolerance $\tau$          & 20$d$         & PDBP, VPSO    \\
  dimension of the problem space        & $n$           & PDBP          \\
  number of neurons in the hidden layer $m$ & 1 .. $n$  & PDBP, VPSO    \\
  dimension of a particle position $d$  & $(n+2)m+1$    & VPSO          \\
  reorder period $T$                    & 10$d$         & VPSO          \\
  population size $N_{pop}$             & 2$d$          & VPSO          \\
  \hline
\end{tabular}
\end{table}
The Sigmoid function is the activation function of an SNN. The SNNs will be evaluated with five KPIs, such as ACC, $F_1$, AUC, TPR and TNR. Two groups of experiments are conducted. The roadmap of experiments is shown in Fig. \ref{fig:roadmap}.

(1) Learnability experiments\\
To examine the learnability of an SNN, obtained by PDBPs and VPSOs, the whole data set is used to train an SNN, in terms of the definition of the learnability of a classifier \cite{Sharma2018}, which is the accuracy obtained by the classifier on the training data. The following experiments are conducted:
\begin{itemize}
\item[i.] An incremental approach is applied to observe the impact of hidden neuron numbers on the performance of a decision maker, by starting from one hidden neuron, increasing a hidden neuron in the hidden layer per step until the number of hidden neurons reaches to the number of attributes for the two data sets.
\item[ii.] The experiments of KPI-PDBPs and KPI-VPSOs on a specific number of hidden neurons are conducted for the two data sets and their performances are compared.
\end{itemize}
(2) Robustness experiments\\
Three experiments are conducted:
\begin{itemize}
    \item[i.] For each data set, ten-fold cross validation experiments are conducted with the specific SNN, obtained by KPI-PDBPs and KPI-VPSOs, respectively.
    \item[ii.] For the two data sets, the experiments of KPI-PDBPs and KPI-VPSOs on different training rates, changing from 0.5 to 0.9, are conducted to examine the impact of training rates on the performance of learned SNNs.
    \item[iii.] Two-fold crossing validation experiments on more benchmark databases that have been used in literature are conducted with the specific SNN, obtained by ACC-PDBP and ACC-VPSO, and the robustness (average performance and standard deviation (denoted as $a\pm b$) is compared with that in literature.
\end{itemize}
*Note: ACC-VPSO indicates that the fitness of VPSO is ACC. Namely ACC will drive the evolution of VPSO. ACC-PDBP denotes the PDBP is driven by ACC.
\subsection{The two benchmark data sets}
The Wisconsin Breast Cancer (WBC) database was created by Wolberg \cite{Wolberg1990}, containing 699 samples, in which 458 samples are benign, and 241 samples are malignant. There are nine basic attributes $x_0, x_1, ..., x_8$, with integer range [1,10]. The missing value of an attribute in an instance of the database is replaced with the mean value of the attribute on the corresponding goal class.

The SMSSpam database \cite{Almeida2011}  has 5574 raw messages, including 747 spams. He et al. \cite{He2016} extracted 20 features from the database, and  the number of features was reduced to 14 by combining some features with similar meanings \cite{He2017}.  We use the 14-attributes database for the experiments. 
\subsection{Learnability Experiments}
\subsubsection{Impact of Hidden Neuron Numbers on the performance of the learned SNNs}
As the experiments in \cite{He2018}, the number of hidden neurons of the SNN is increased from 1 to $n$, where $n$ is the input number of the problem space.

\begin{figure}[htp]
\subfigure[The KPIs of different SNNs, obtained by ERR-PDBP]
{\begin{minipage}{0.49\textwidth}
\includegraphics[height=1.6in]{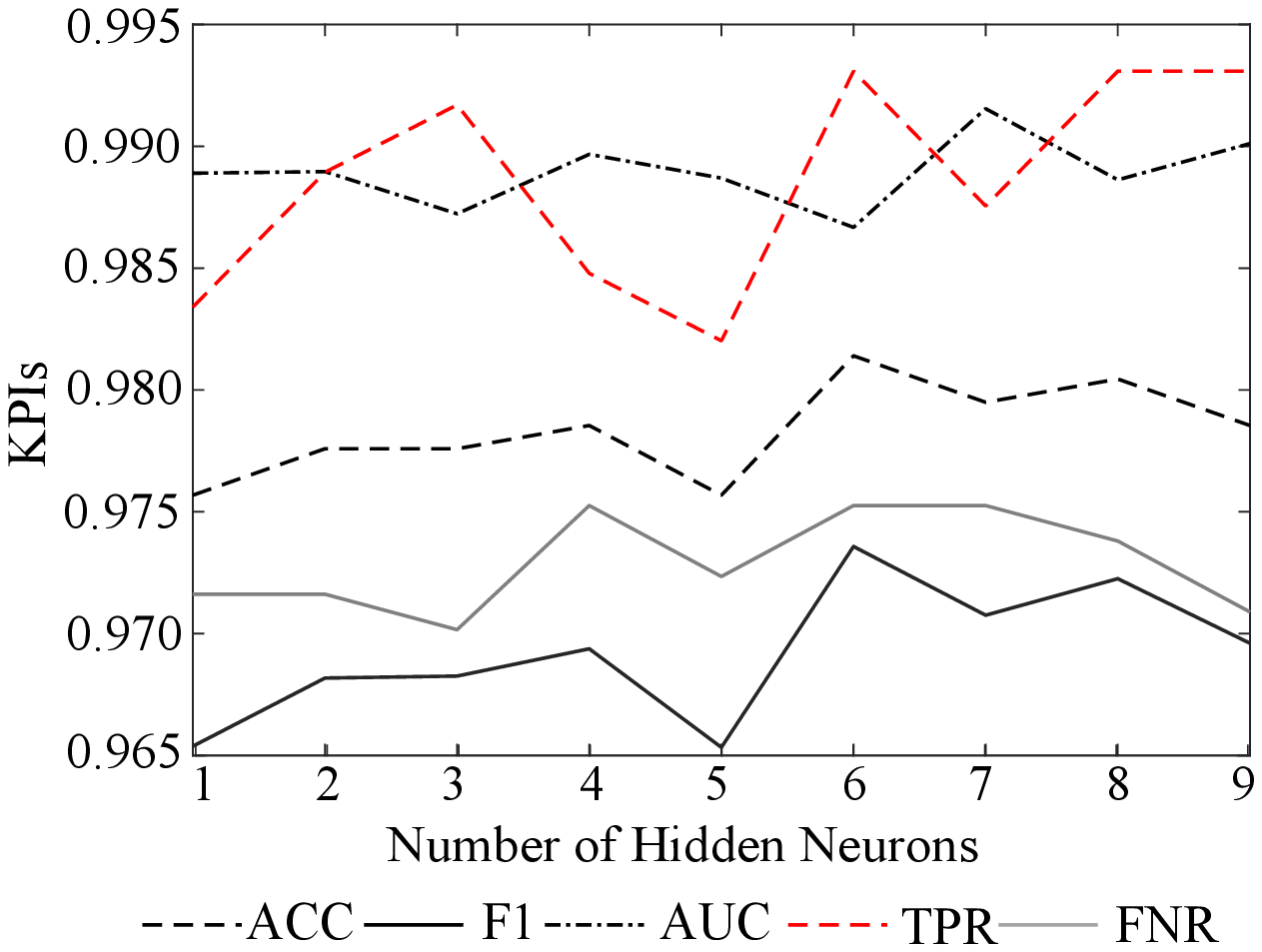}
\end{minipage}}
\subfigure[The KPIs of different SNNs, obtained by ERR-VPSO]
{\begin{minipage}{0.49\textwidth}
\includegraphics[height=1.6in]{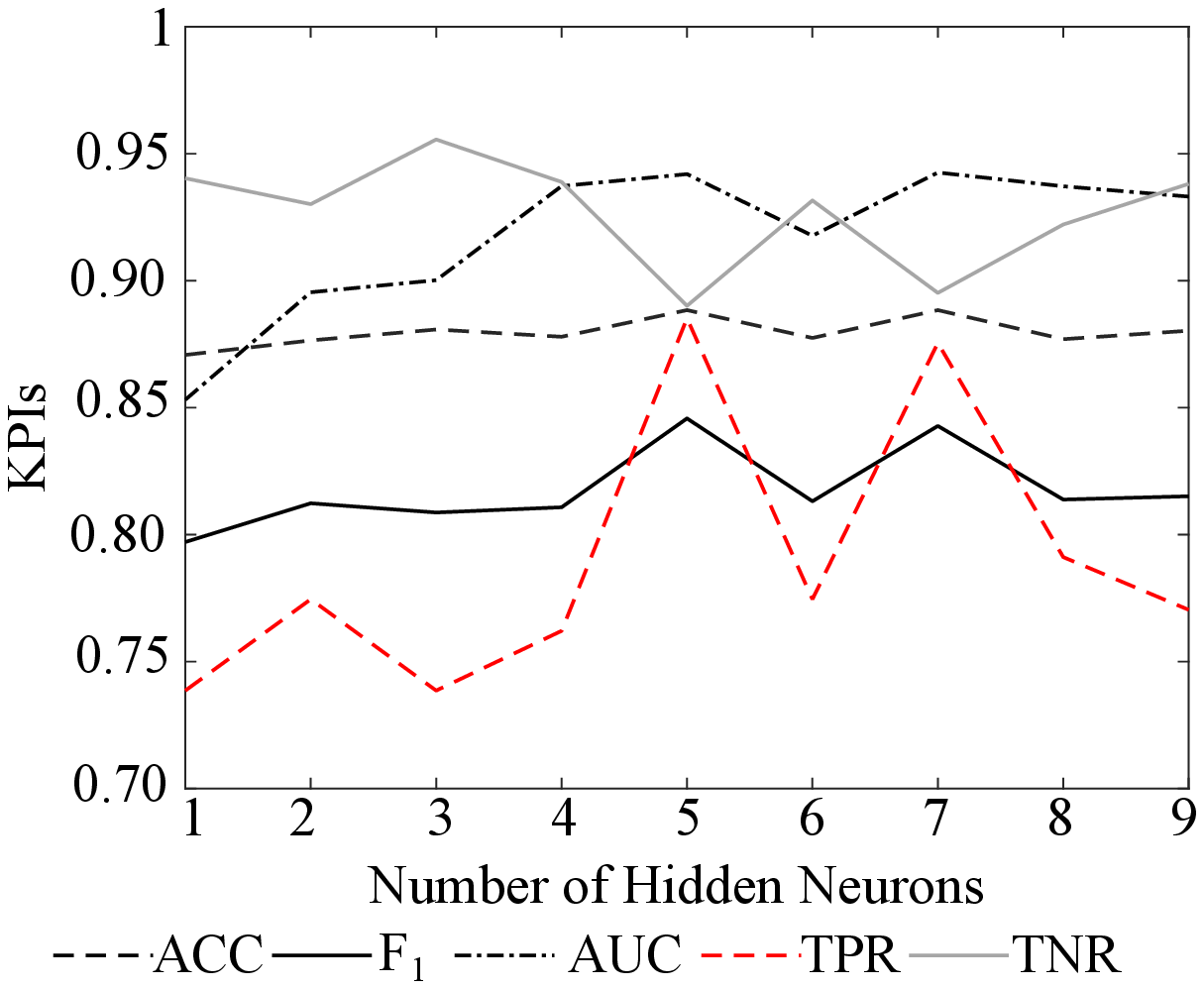}
\end{minipage}}
\subfigure[The KPIs of different SNNs, obtained by ACC-PDBP]
{\begin{minipage}{0.49\textwidth}
\includegraphics[height=1.6in]{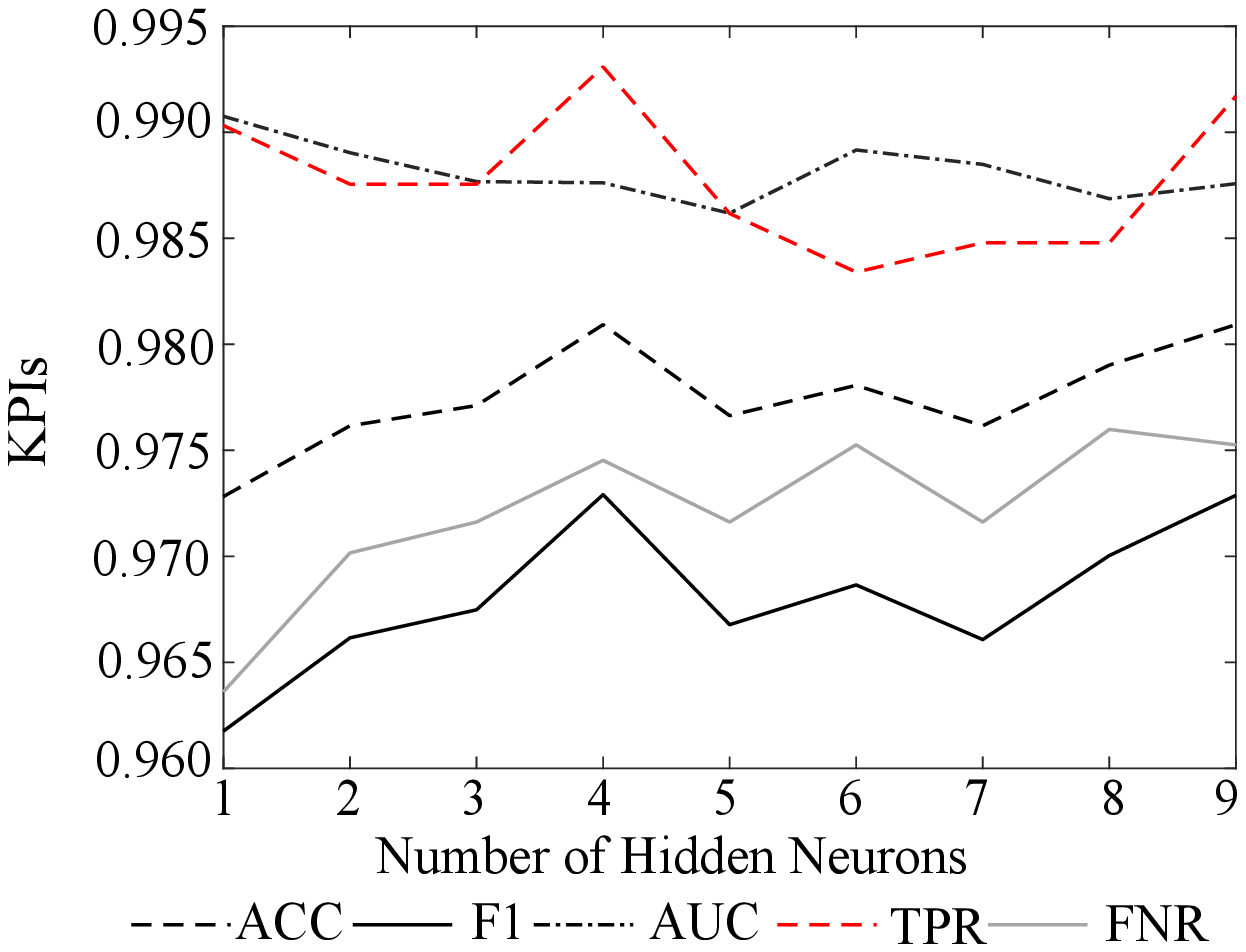}
\end{minipage}}
\subfigure[The KPIs of different SNNs, obtained by ACC-VPSO]
{\begin{minipage}{0.49\textwidth}
\includegraphics[height=1.6in]{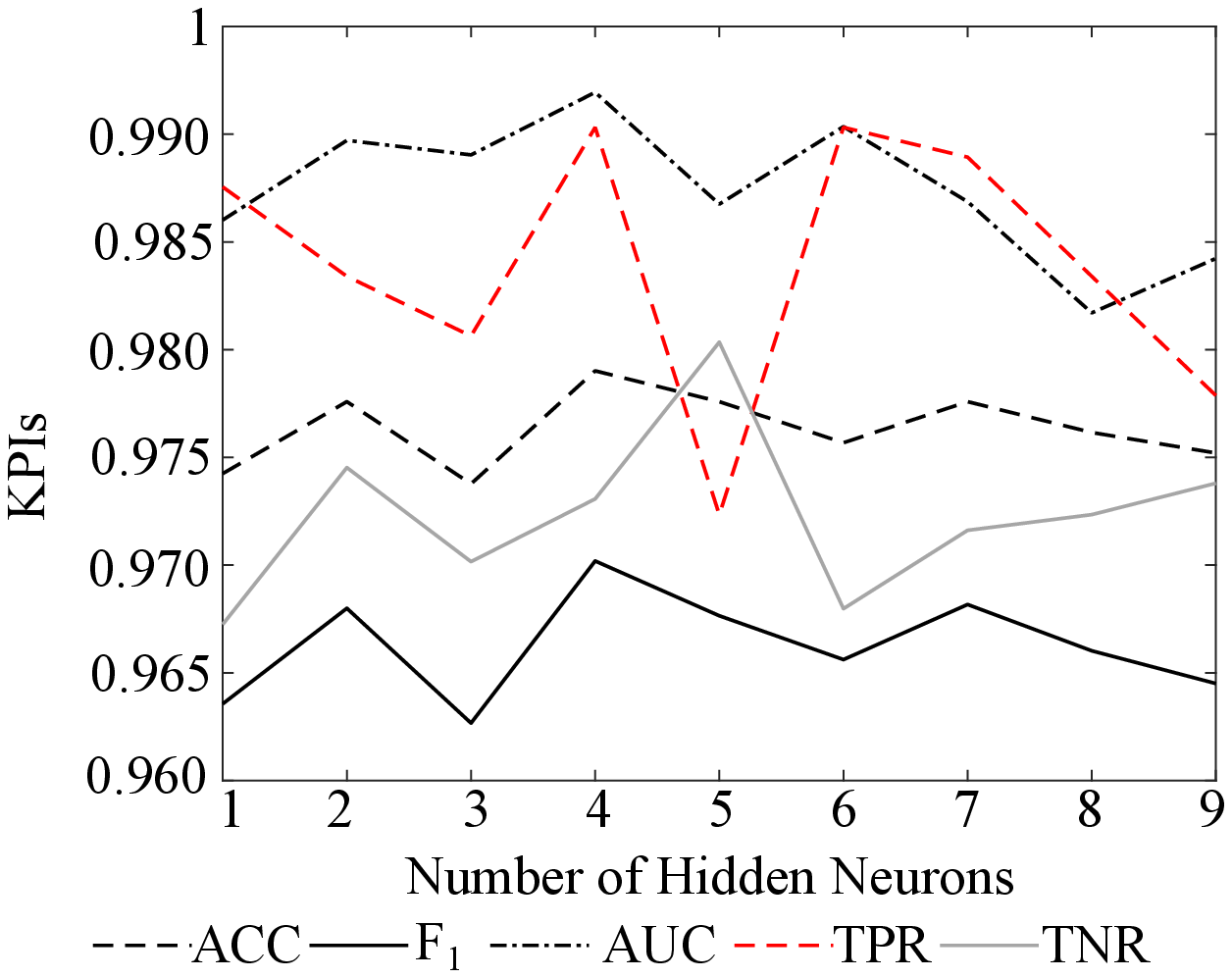}
\end{minipage}}
\subfigure[The KPIs of different SNNs, obtained by $F_1$-PDBP]
{\begin{minipage}{0.49\textwidth}
\includegraphics[height=1.6in]{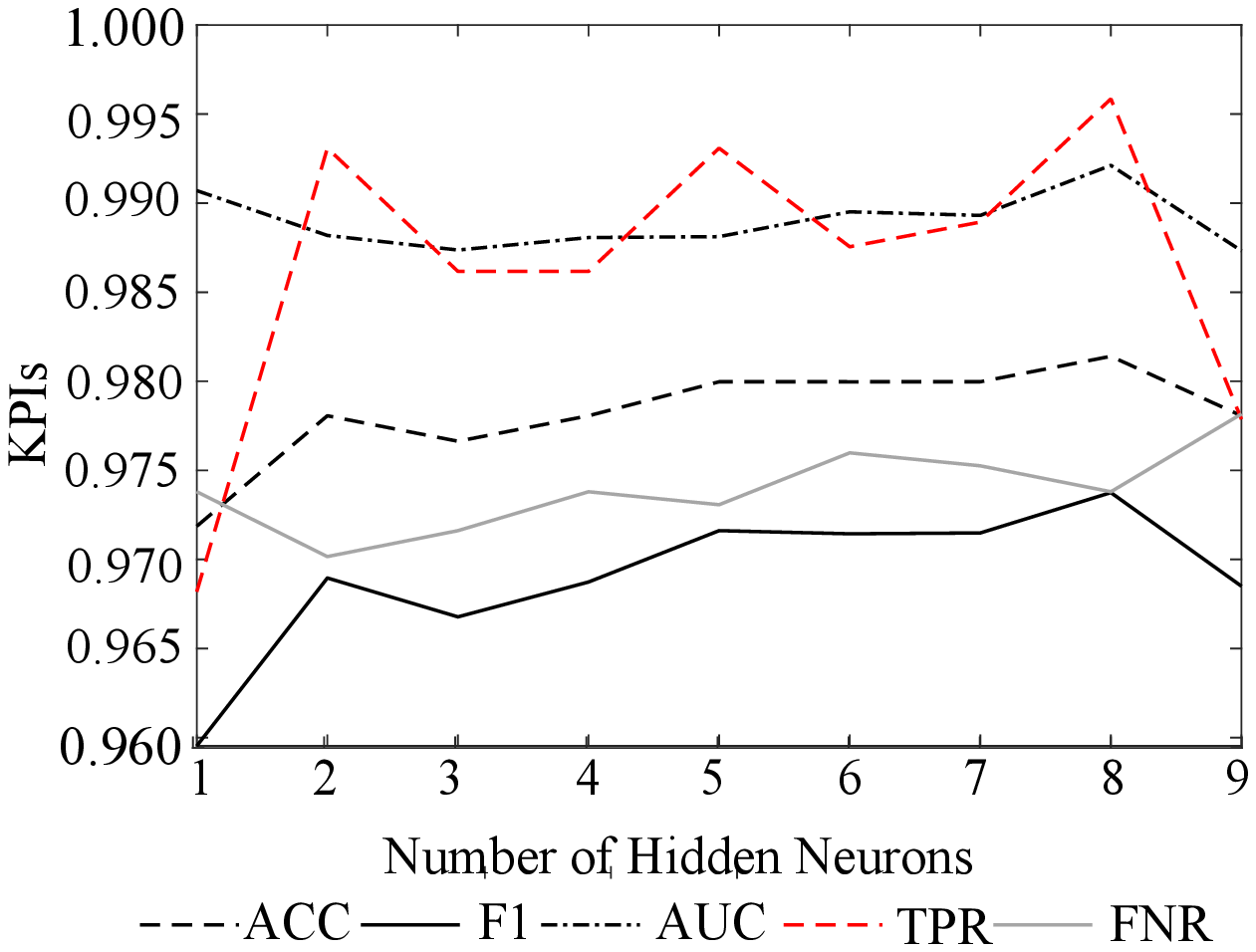}
\end{minipage}}
\subfigure[The KPIs of different SNNs, obtained by $F_1$-VPSO]
{\begin{minipage}{0.49\textwidth}
\includegraphics[height=1.6in]{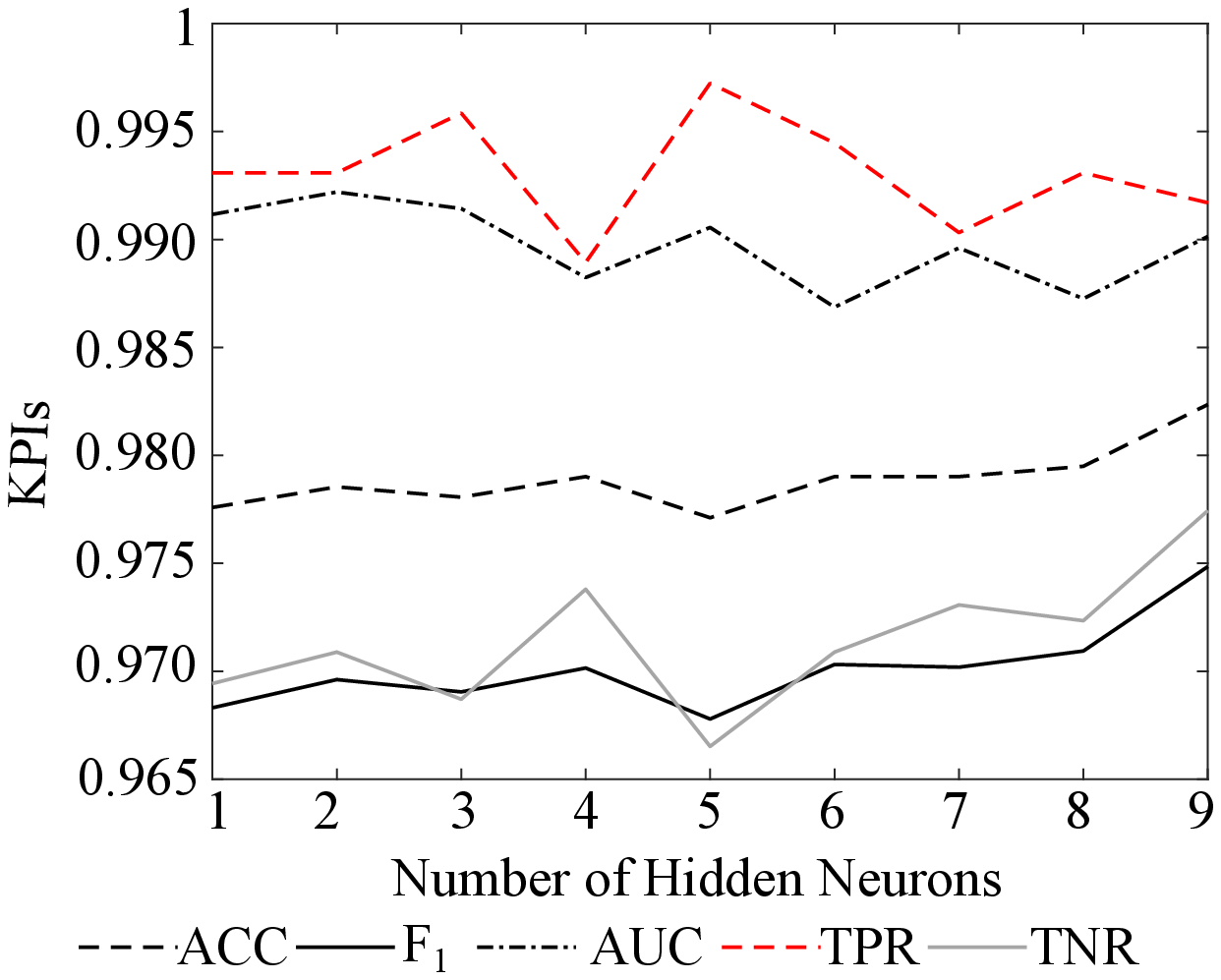}
\end{minipage}}
\caption{The performance of SNNs with different numbers of hidden neurons on WBC, obtained by KPI-PDBPs and KPI-VPSOs, where KPIs= \{ERR, ACC, $F_1$\}.} \label{fig:Impact-SNN-WBC}
\end{figure}

\paragraph{\bf On the WBC database\\}
Figs. \ref{fig:Impact-SNN-WBC}(a)-(f) show the performances of SNNs on WBC, obtained by KPI-PDBPs and KPI-VPSOs. It can be seen that TPR and AUC are on the top alternatively and $F_1$ < TNR < ACC in those figures produced by KPI-PDBPs and ACC-VPSO.

For VPSO, different KPIs could have different impact on the performances of the optimised SNNs.  As shown in \ref{fig:Impact-SNN-WBC} (b), produced by ERR-VPSO, AUC and TNR are on the top alternatively, while TPN is at the bottom. This might be because that the number of negatives in the data set is larger than the number of positives, ERR, as the fitness of ERR-VPSO, makes the search bear to improving TNR. In contrast, in \ref{fig:Impact-SNN-WBC} (f), produced by $F_1$-VPSO, TPR is on the top, while TNR and $F_1$ are at the bottom closely. This might be because that $F_1$, as the fitness of $F_1$-VPSO makes the search bear to improving TPR.

\begin{figure}[htp]
\subfigure[The KPIs of different SNNs, obtained by ERR-PDBP]
{\begin{minipage}{0.49\textwidth}
\includegraphics[height=1.6in]{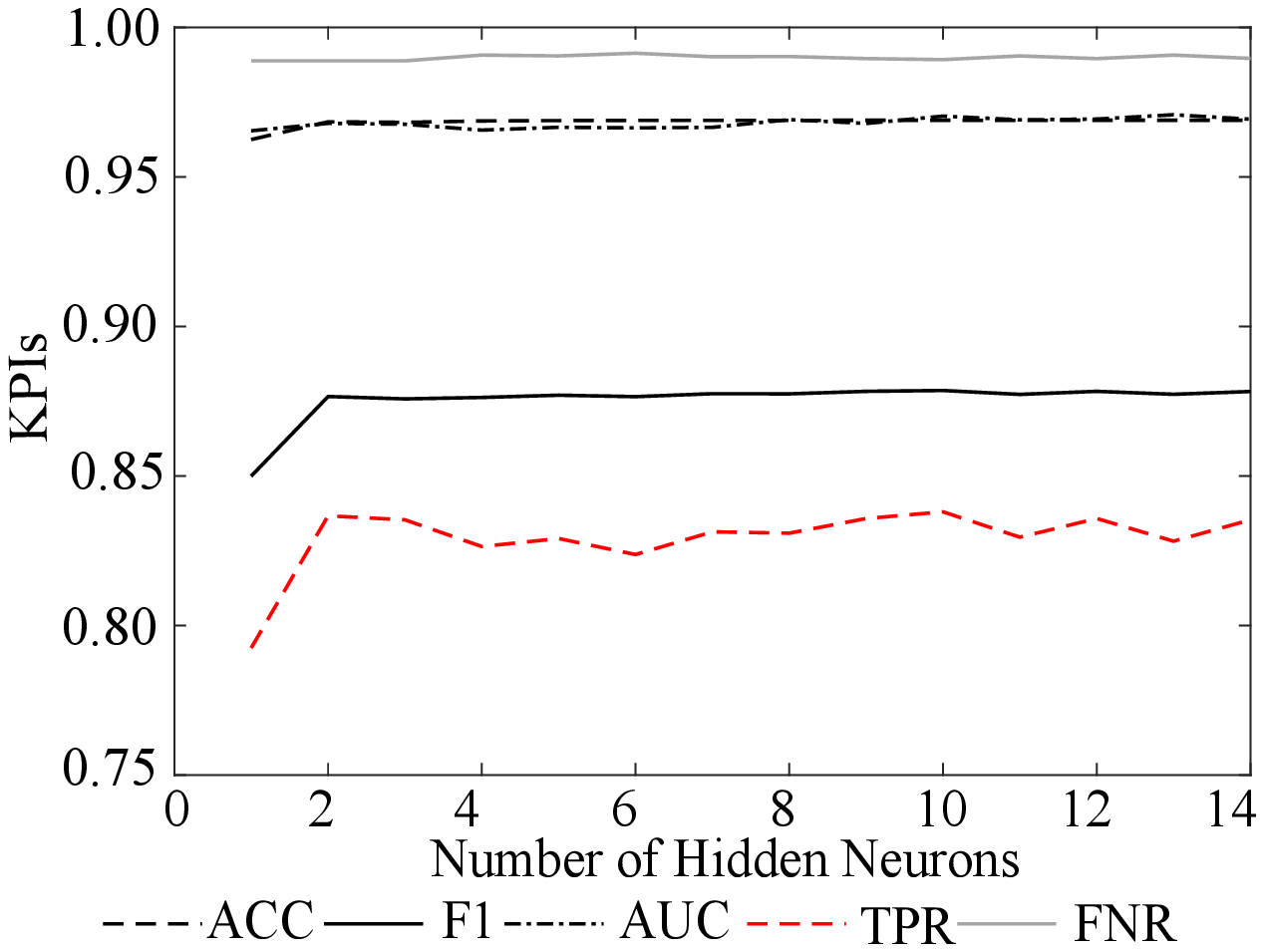}
\end{minipage}}
\subfigure[The KPIs of different SNNs, obtained by ERR-VPSO]
{\begin{minipage}{0.49\textwidth}
\includegraphics[height=1.6in]{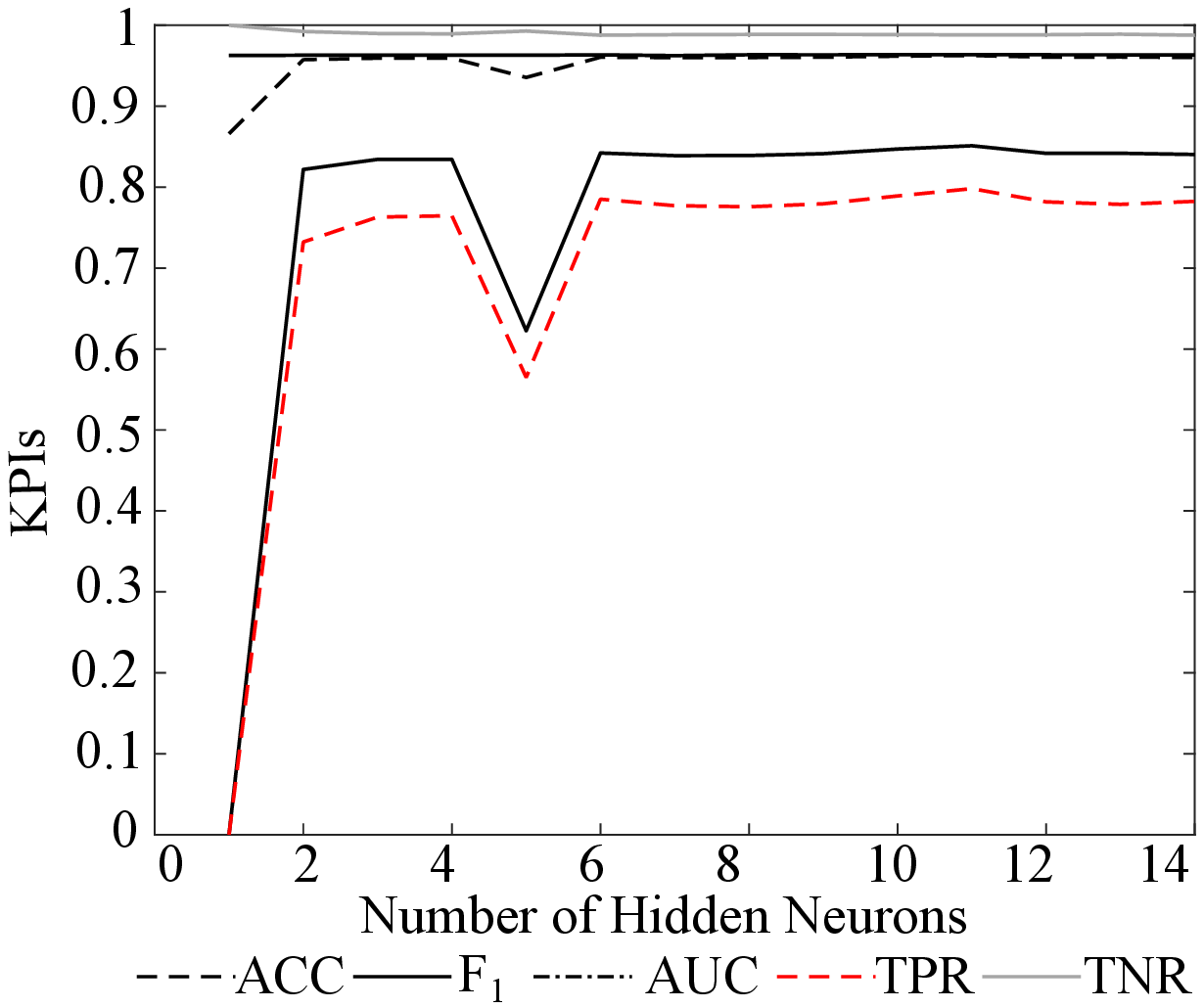}
\end{minipage}}
\subfigure[The KPIs of different SNNs, obtained by ACC-PDBP]
{\begin{minipage}{0.49\textwidth}
\includegraphics[height=1.6in]{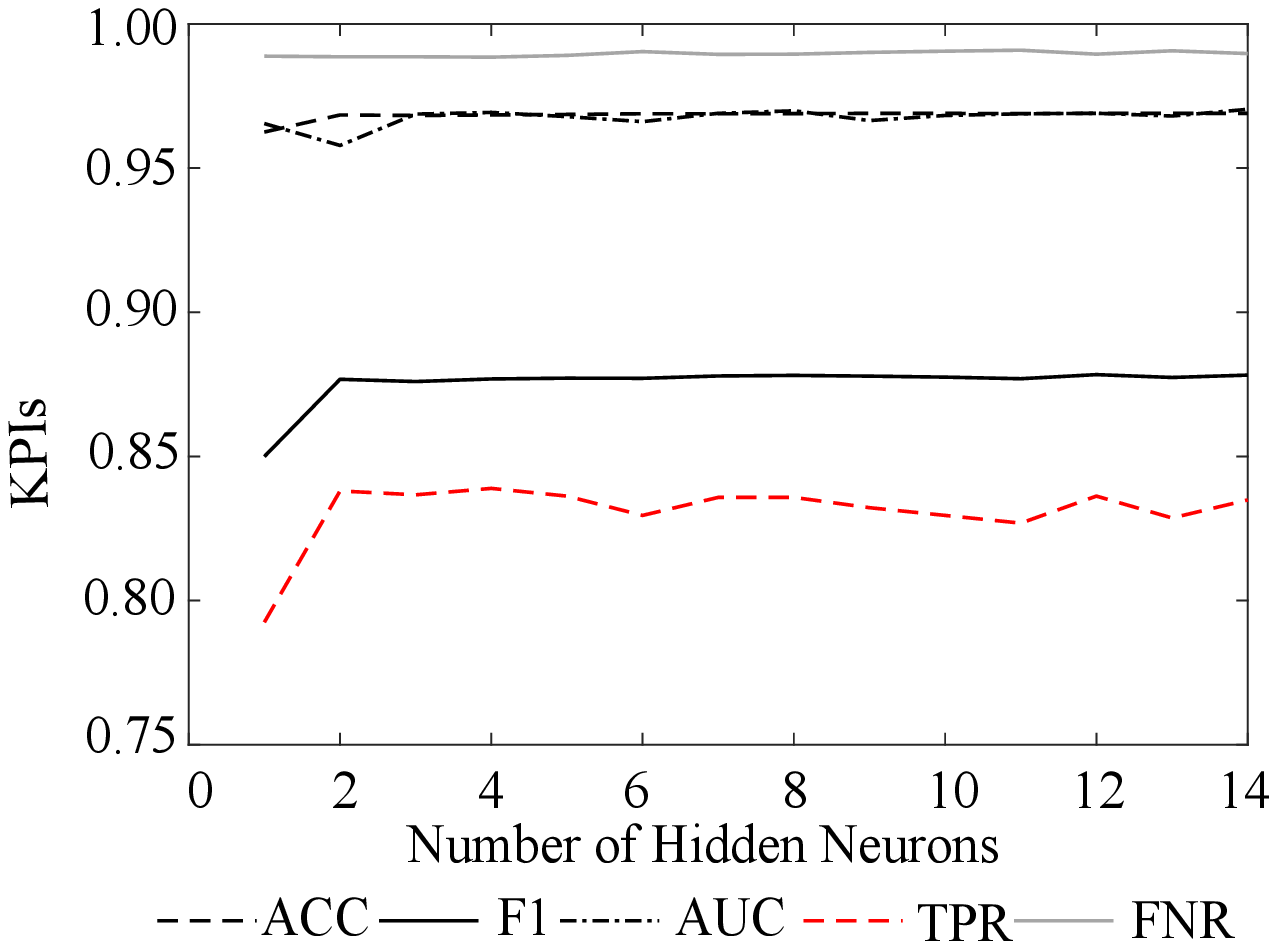}
\end{minipage}}
\subfigure[The KPIs of different SNNs, obtained by ACC-VPSO]
{\begin{minipage}{0.49\textwidth}
\includegraphics[height=1.6in]{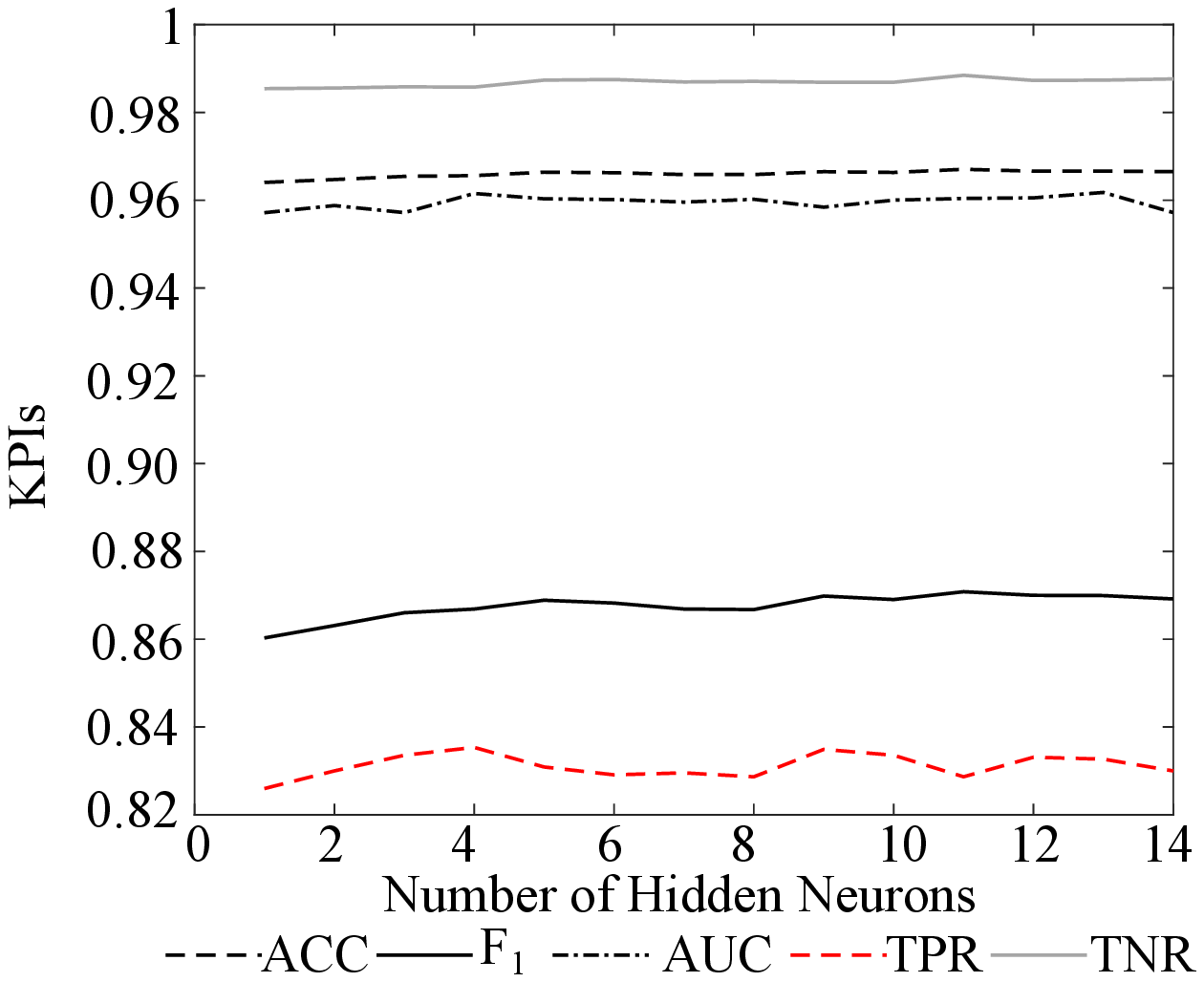}
\end{minipage}}
\subfigure[The KPIs of different SNNs, obtained by $F_1$-PDBP]
{\begin{minipage}{0.49\textwidth}
\includegraphics[height=1.6in]{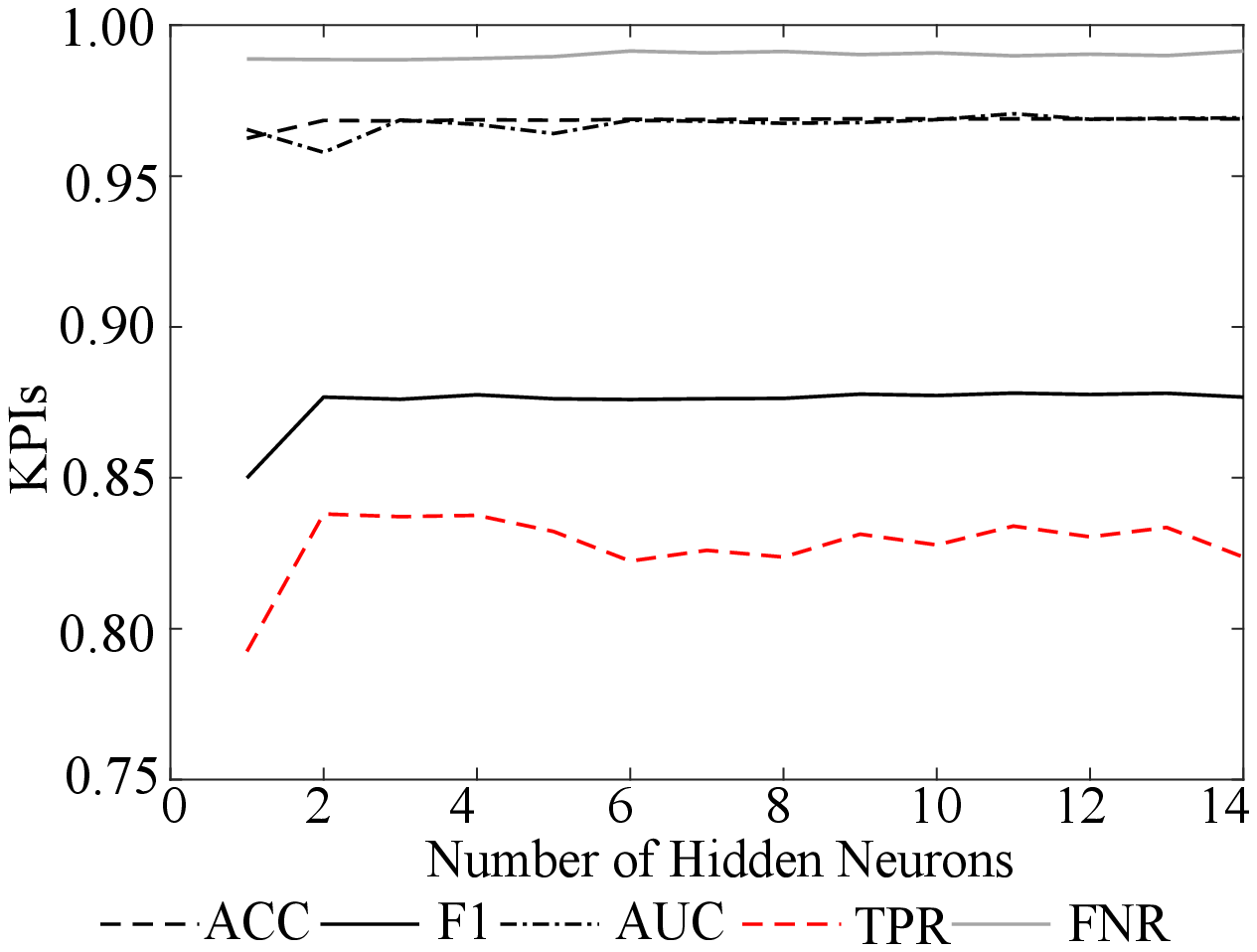}
\end{minipage}}
\subfigure[The KPIs of different SNNs, obtained by $F_1$-VPSO]
{\begin{minipage}{0.49\textwidth}
\includegraphics[height=1.6in]{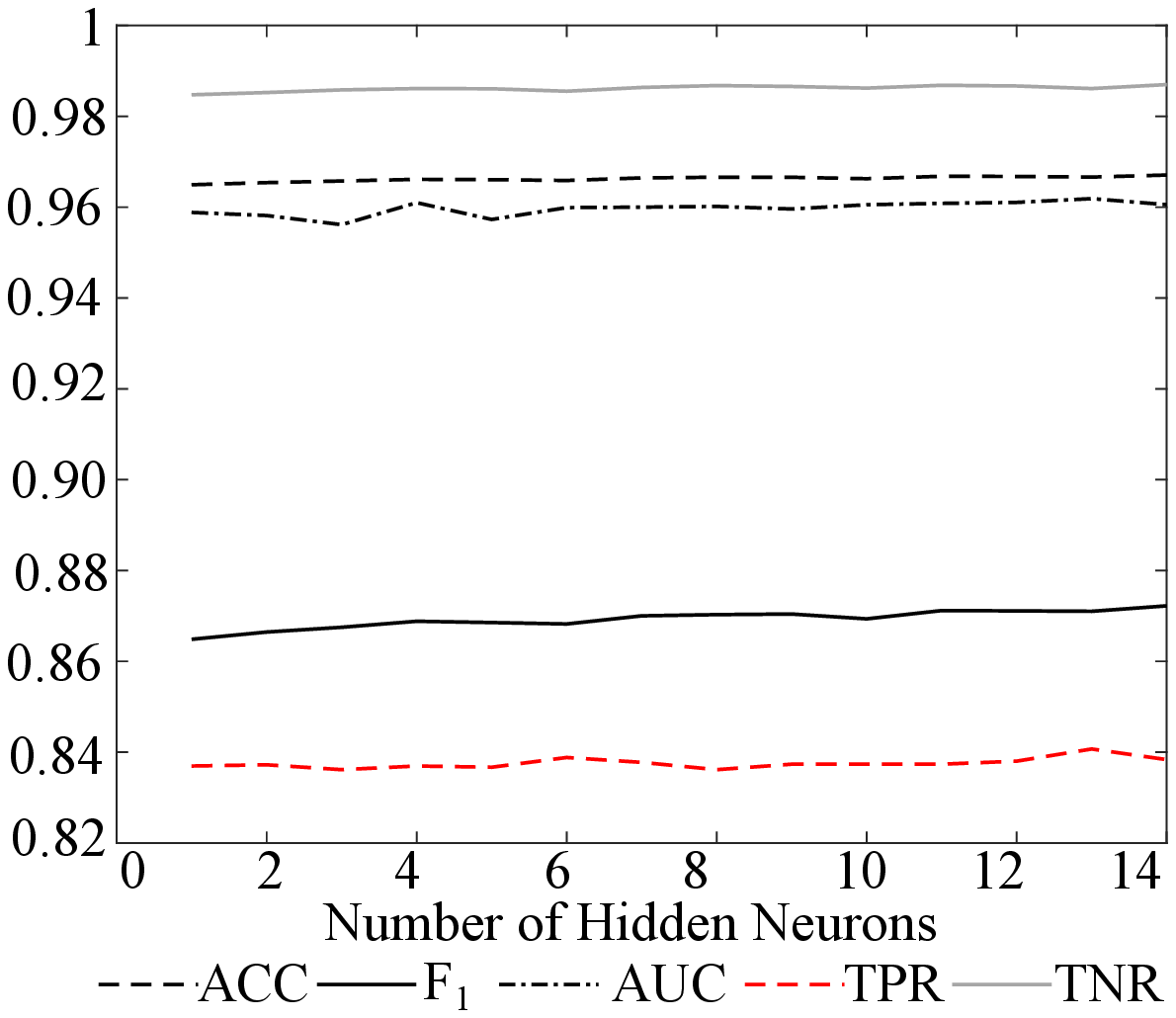}
\end{minipage}}
\caption{The performance of SNNs with different numbers of hidden neurons on the SMS Spam database, obtained by KPI-PDBP and KPI-VPSO, KPI$\in$\{ERR, ACC, $F_1$\}} \label{fig:Impact-SNNs-SMS}
\end{figure}

From Figs. \ref{fig:Impact-SNN-WBC} (a) - (f), when the number of hidden neurons reaches 4, the performance has reached at a certain level, after then, the curves of KPIs  fluctuate in a certain range. Therefore, the following experiments are particularly examine the performance of the SNN with 4 hidden neurons (denoted as SNN$_4$), obtained by KPI-PDBP and KPI-VPSO on WBC. The fluctuation of performances are large. This indicates that the learning factor of PDBPs and VPSOs should be smaller.
\paragraph{\bf On the SMS Spam database\\}
Figs. \ref{fig:Impact-SNNs-SMS}(a)-(f) show the performances of SNNs on the SMS database, obtained by KPI-PDBPs and KPI-VPSOs, where KPIs=\{ERR, ACC, $F_1$\}.
It can be seen that the values of the five KPIs have the same order: TPR < $F_1$ < AUC $\approx$ ACC < TNR in all figures. TPR has a small fluctuation. The SMS data set is a unbalanced data set with only 747 positives in 5574 samples. The tiny change on the estimates of negatives cannot produce obvious change of TNR and other KPIs, but the tiny change on the estimates of positives could be shown on TPR.

From Figs. \ref{fig:Impact-SNNs-SMS}(a)-(c) and (e), produced by KPI-PDBPs and ERR-VPSO, it can be seen that when the number of hidden neurons is larger than 1, the performances almost keep at the same level. However, the SNN with only one neuron, obtained by ACC-VPSO and $F_1$-VPSO, can get the same level of performances as the SNNs with more hidden neurons.

This might tell us that the performances of SNNs cannot be improved further, unless further information or new features could be provided.

\subsubsection{Experiments for a specific SNN on each data set}
\begin{figure}[htp]
\centering
\subfigure[Error evolution process of the SNN$_4$]
{\begin{minipage}{0.49\textwidth}
\includegraphics[height=1.8in]{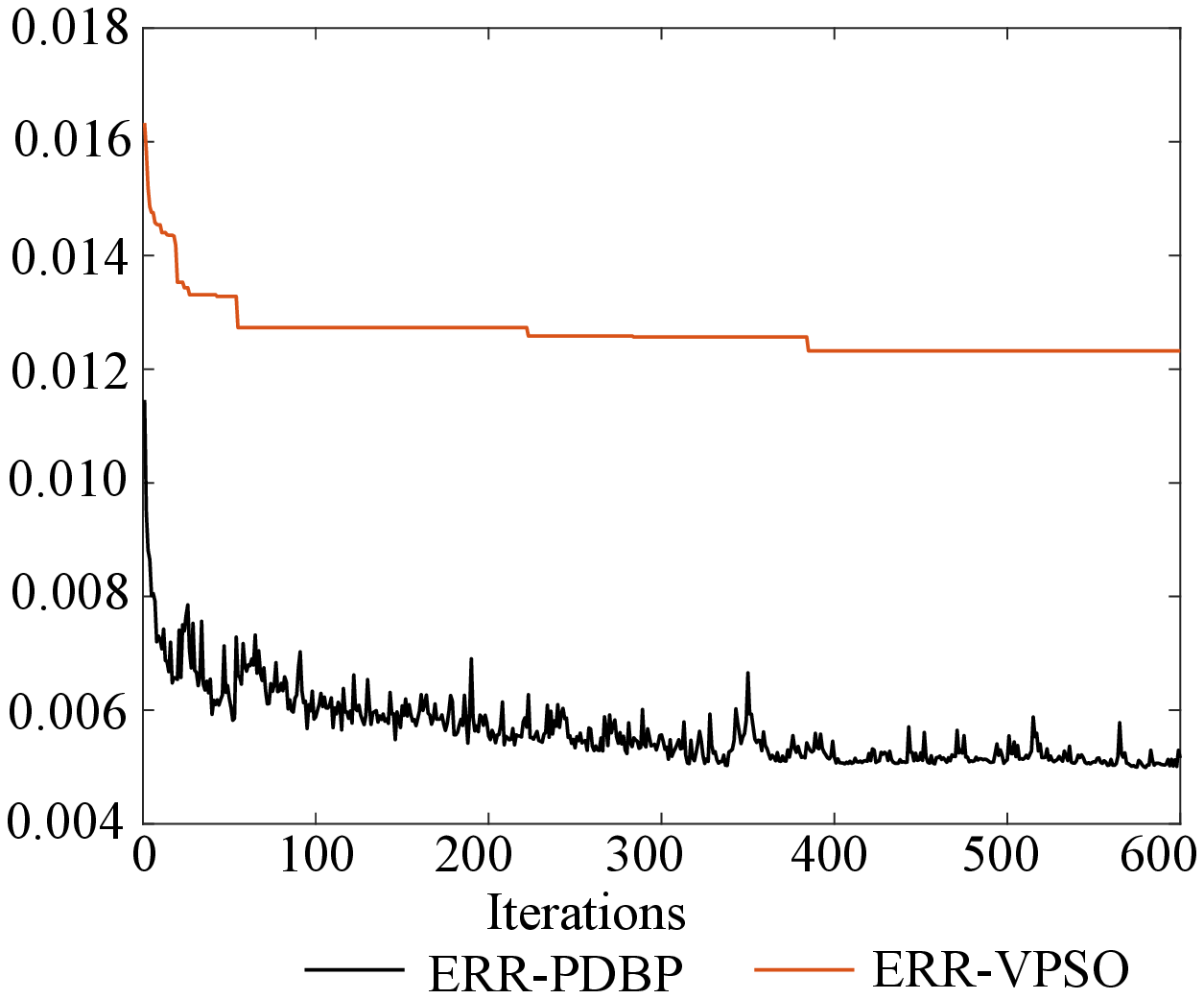}
\end{minipage}}
\subfigure[ACC evolution process of the SNN$_4$]
{\begin{minipage}{0.49\textwidth}
\includegraphics[height=1.8in]{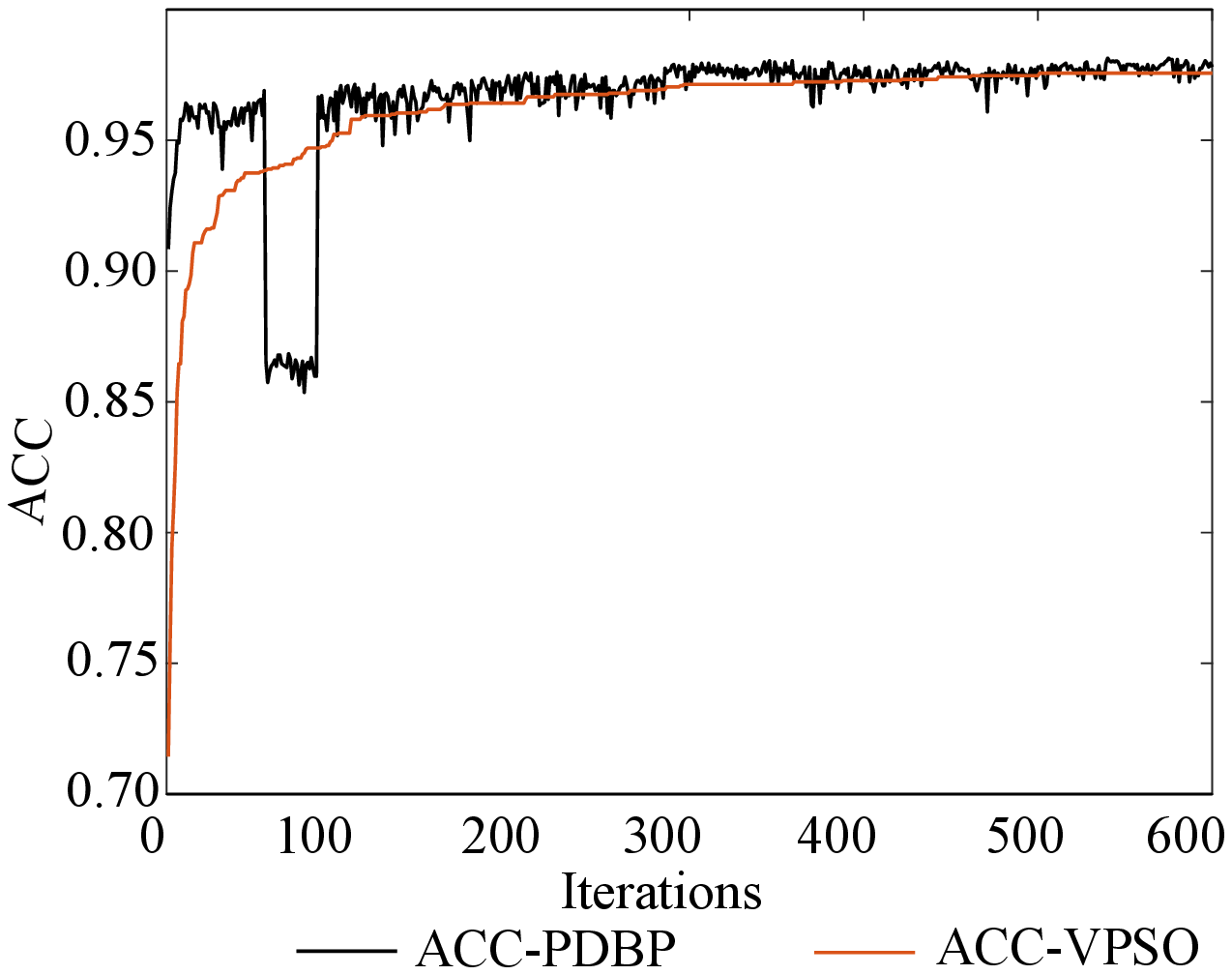}
\end{minipage}}
\subfigure[$F_1$ evolution process of the SNN$_4$]
{\begin{minipage}{0.49\textwidth}
\includegraphics[height=1.8in]{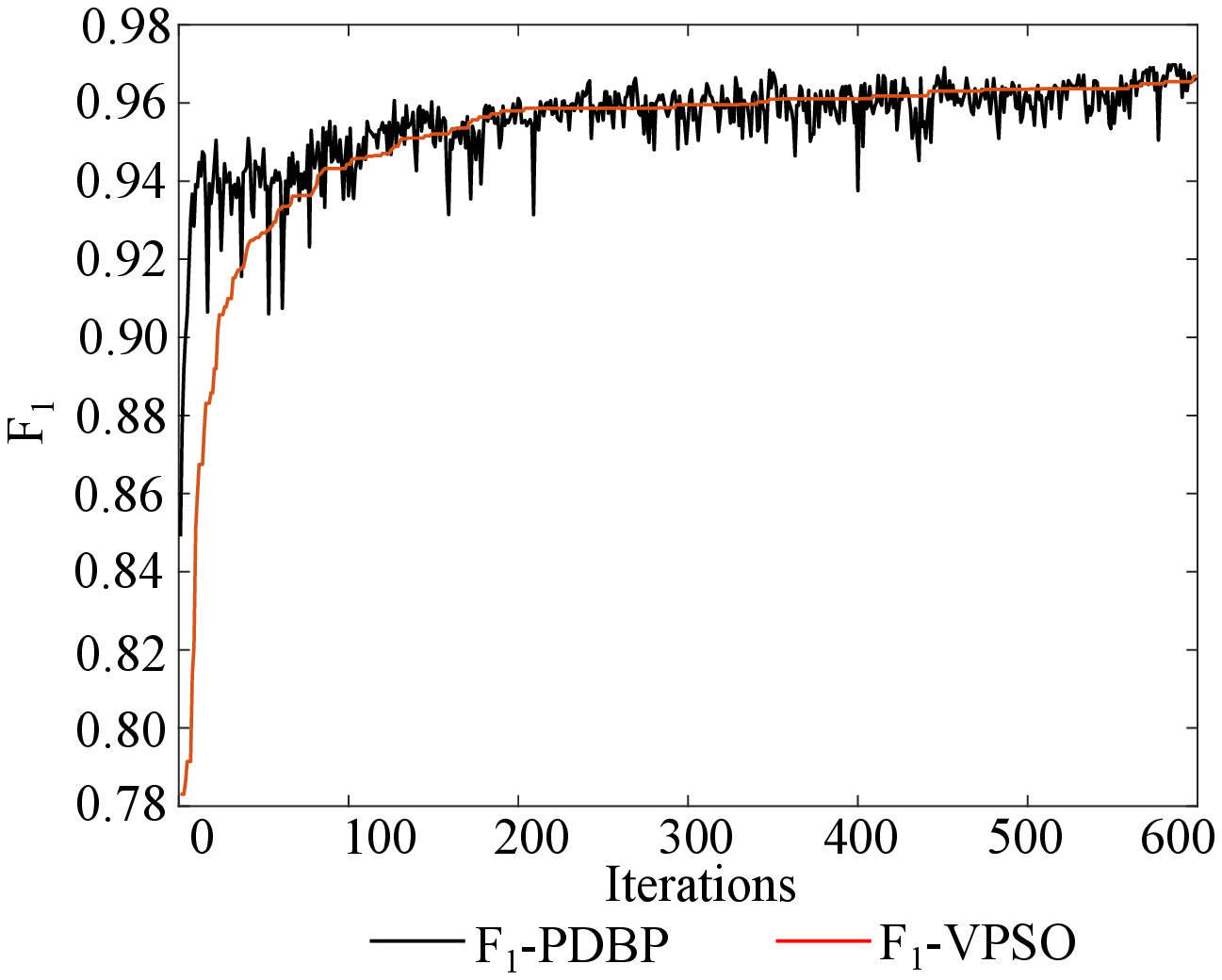}
\end{minipage}}
\caption{Performance evolutionary process of SNN$_4$, searched by PDBP and VPSO with ERR, ACC and $F_1$ as the fitness, respectively} \label{fig:evol-WBC}
\end{figure}

\paragraph{\bf Evolutionary process of PDBPs and VPSOs on WBC\\}
The evolutionary process of SNN$_4$ is examined. For readability, the first 600 iterations are displayed in figures. Figs. \ref{fig:evol-WBC} (a)-(c) show the evolutionary process of errors, searched by KPI-PDBPs and KPI-VPSOs on WBC, respectively.

Fig. \ref{fig:evol-WBC} (a) shows the error evolutionary process of SNN$_4$, searched by ERR-PDBP and ERR-VPSO. It can been seen that the errors, evolved by ERR-VPSO, are much larger than those, evolved by ERR-PDBP. However, from Fig. \ref{fig:evol-WBC} (b), it can been seen that the accuracy evolutionary processes for ACC-PDBP and ACC-VPSO are similar, but before generation 100, there is a gap of drops for ACC-PDBP.  From Fig. \ref{fig:evol-WBC} (c)  the $F_1$ evolutionary processes of $F_1$-PDBP and $F_1$-VPSO are very close.

\begin{figure}[htp]
\centering
\subfigure[Error evolution process of the SNN$_7$]
{\begin{minipage}{0.49\textwidth}
\includegraphics[height=1.8in]{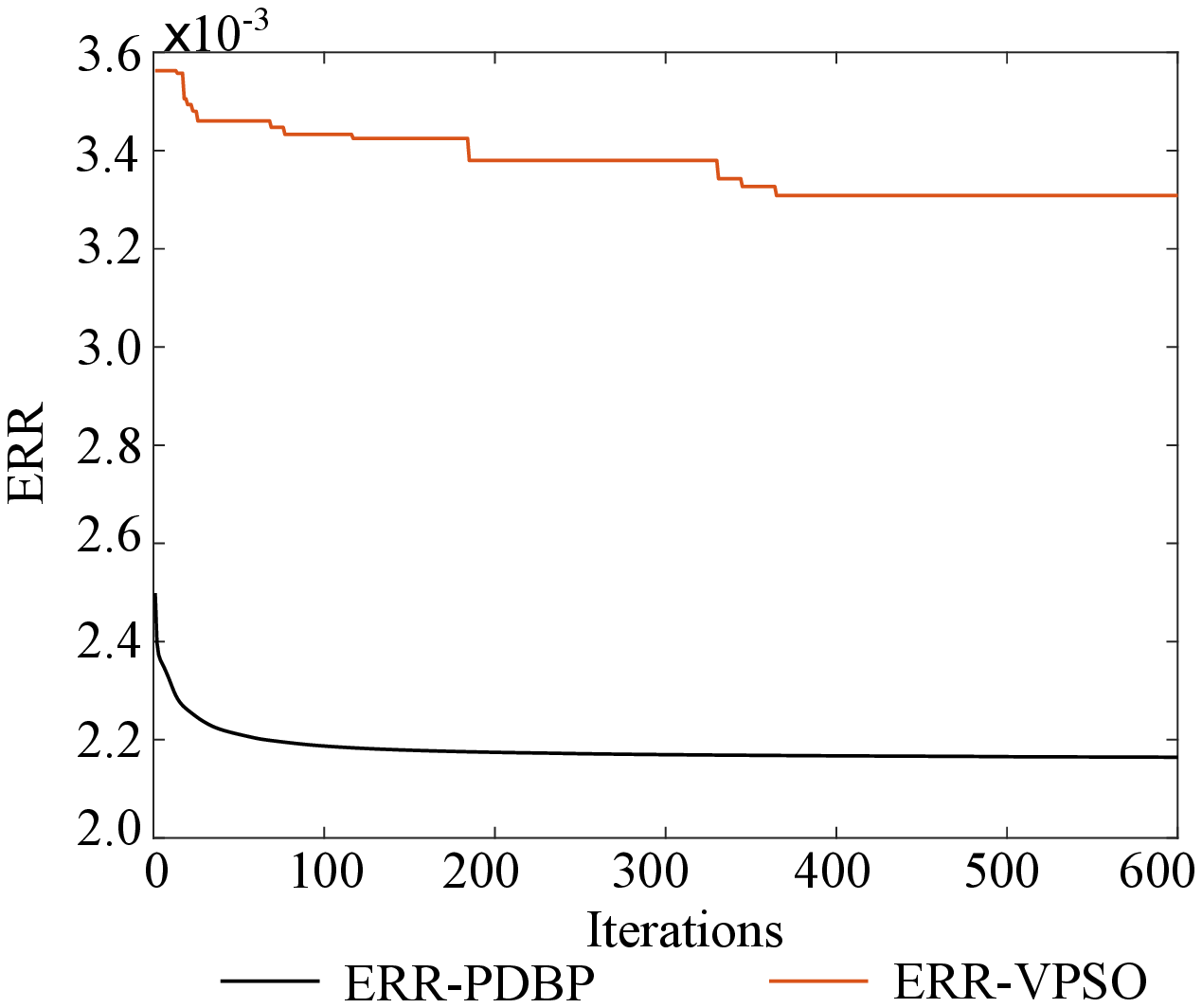}
\end{minipage}}
\subfigure[ACC evolution process of the SNN$_7$]
{\begin{minipage}{0.49\textwidth}
\includegraphics[height=1.8in]{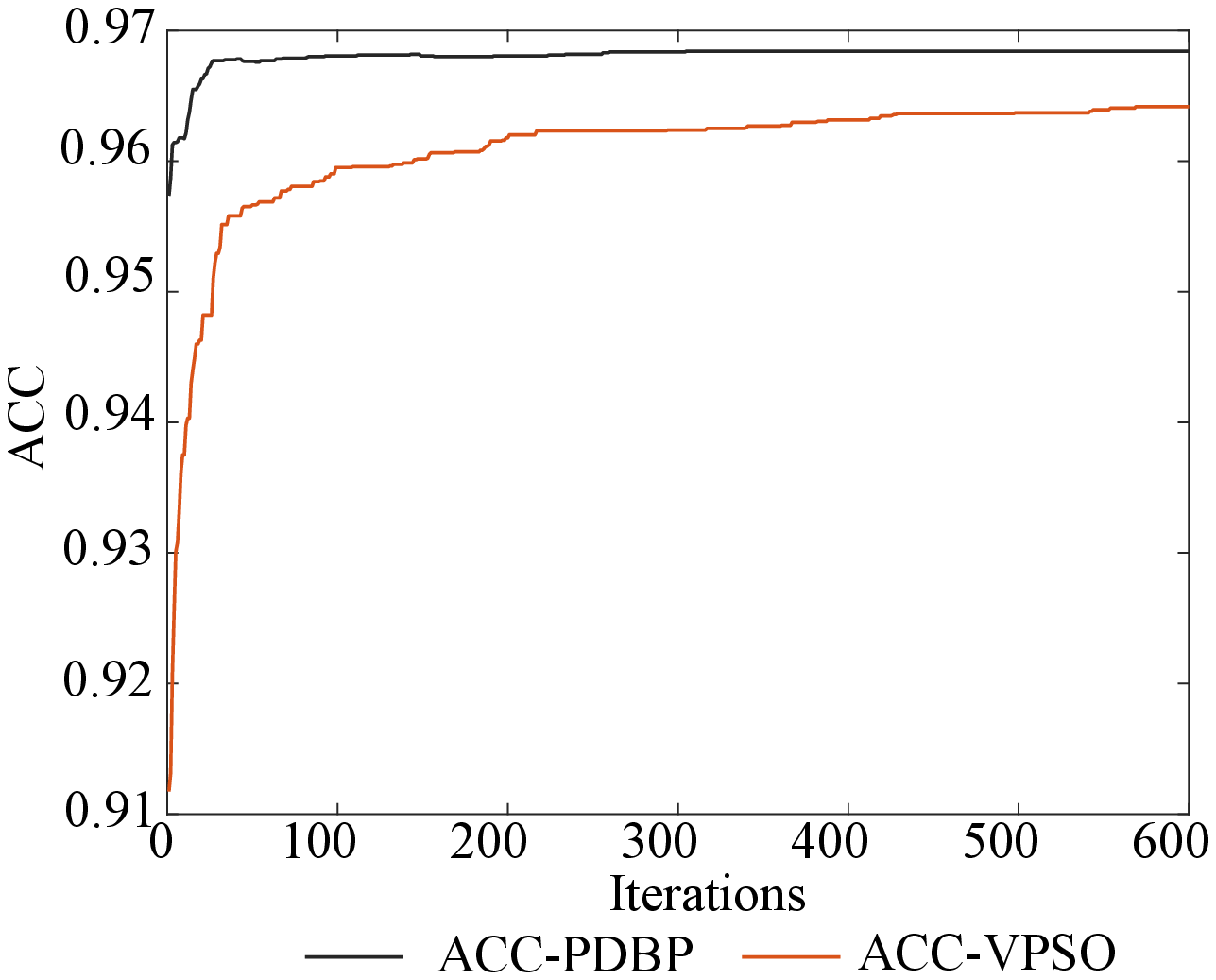}
\end{minipage}}
\subfigure[$F_1$ evolution process of the SNN$_7$]
{\begin{minipage}{0.49\textwidth}
\includegraphics[height=1.8in]{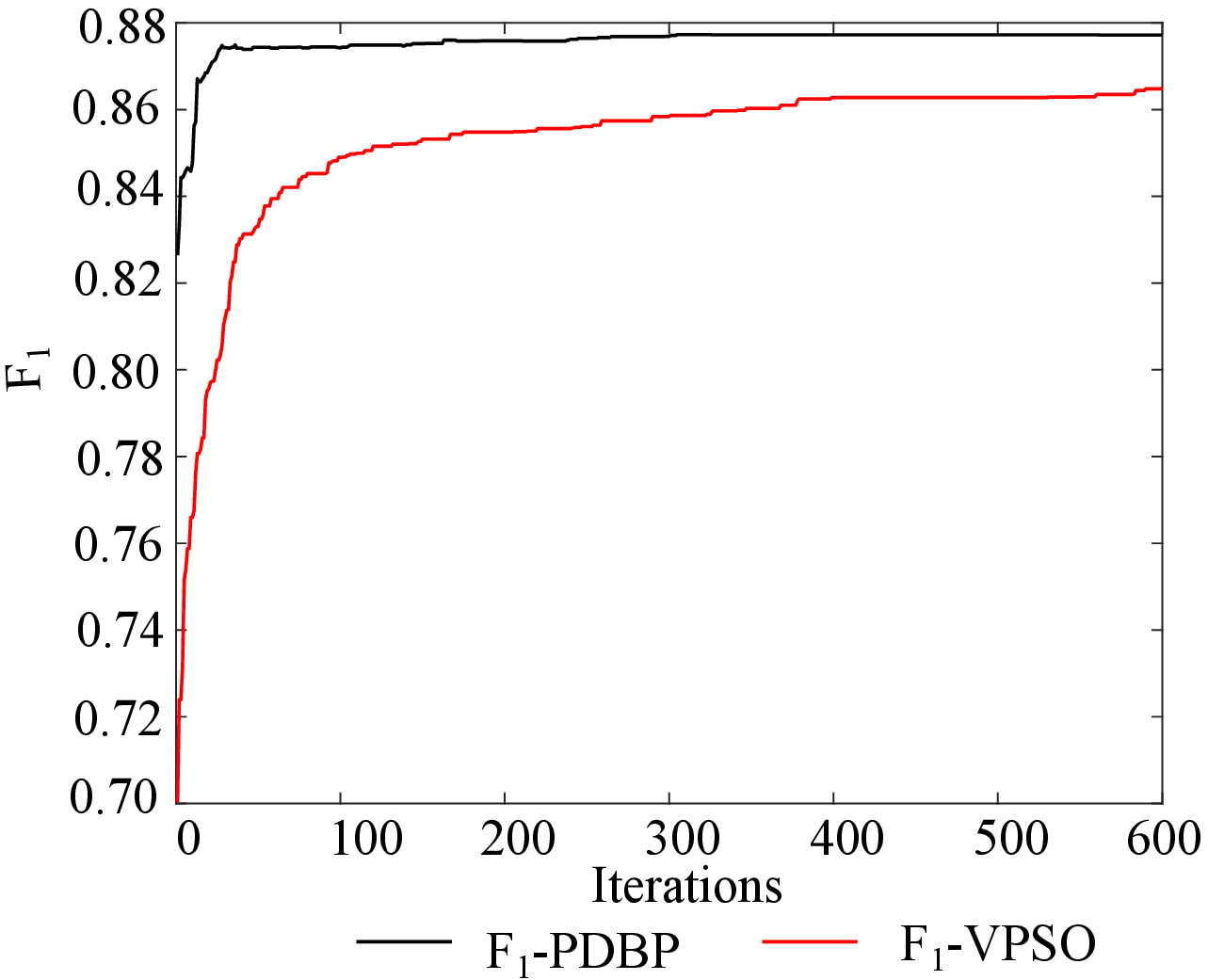}
\end{minipage}}
\caption{Performance evolutionary process of the SNN$_7$, obtained by PDBP and VPSO} \label{fig:evol-SMS}
\end{figure}

However, Figs. \ref{fig:evol-WBC} (a)-(c), the evolutionary processes of KPI-PDBPs fluctuate very much. This further indicates that the learning factor of SNNs could be smaller for KPI-PDBPs. For KPI-VPSOs, as the global best solution is recorded in each iteration, the fluctuation cannot be seen.
\paragraph{\bf Evolutionary processes of KPI-PDBPs and KPI-VPSOs on SMS\\}
The evolutionary process of SNN$_7$ is examined. In the same way, 600 iterations are displayed in figures. Figs. \ref{fig:evol-SMS} (a)-(c) show the evolutionary process of ERR, ACC and $F_1$, searched by different KPI-PDBPs and KPI-VPSOs on SMS, respectively.

Fig. \ref{fig:evol-SMS} (a) shows the $ERR$ evolutionary process of SNN$_7$, searched by ERR-PDBP and ERR-VPSO. Similar to the case on WBC.  It can been seen that the errors, evolved by ERR-VPSO, are much larger than those, evolved by ERR-PDBP. But different to the case on WBC, from Fig. \ref{fig:evol-SMS} (b), it can been seen that the values of ACC, evolved by ACC-PDBP are better than those, evolved by ACC-VPSO. Also, from Fig. \ref{fig:evol-SMS} (c), the values of $F_1$, evolved by $F_1$-PDBP are better than that, evolved by $F_1$-VPSO.

Compared with the curves obtained by PDBPs on WBC in Figs. \ref{fig:evol-WBC} (a)-(c), the curves, obtained by KPI-PDBPs on SMS in Figs. \ref{fig:evol-SMS} (a)-(c), are much smoother. This indicates that the learning factor of KPI-PDBP on SMS is appropriate.
\paragraph{\bf Performances of the SNN$_4$ obtained by KPI-PDBPs and KPI-VPSOs on WBC\\}
Table \ref{tab:KPIs-SNN4-WBC} shows the five KPIs of SNN$_4$, obtained by PDBPs an VPSOs, driven by the three KPIs, on WBC, the best iteration index (I) when the specific KPI has not changed for $\tau$ (i.e. $20d$) iterations, as well as the training time (T) in the unit of millisecond.

Comparing the KPIs of SNN$_4$ obtained by ERR-PDBP and ERR-VPSO in Table \ref{tab:KPIs-SNN4-WBC}, ERR-PDBP obtains much better KPIs than ERR-VPSO. Although the iteration number of ERR-PDBP is much larger than that of ERR-VPSO, the training time of ERR-PDBP is still much shorter than that of ERR-VPSO. Assume we simply divide each step of PDBP to two tasks: Forward propagation (FP) and Back propagation (BP), the time of FP is $O(Nd)$ and the time of BP is $O(Nd)$. Therefore, the time complexity of PDBP is $O(INd)$.   For VPSO, each individual will have the computing task of FP without the BP  computing task, the size of population is 2d, the computing complexity of VPSO is $O(INd^2)$.That is why the training time of VPSO is much larger than PBDP, even the iteration number of VPSO is smaller than that of PDBP.

Comparing the performances of ACC-PDBP and ACC-VPSO, it can be seen that ACC-PDBP obtains better performed SNNs in ACC, $F_1$, TPR and TNR, whereas ACC-VPSO obtains better performed SNNs in AUC only. The iteration number and training time of ACC-PDBP is smaller than that of ACC-VPSO.

Comparing the performances of SNNs obtained by $F_1$-PDBP and $F_1$-VPSO, the $F_1$-VPSO obtains better performed SNNs in ACC, $F_1$, AUC and TPR, while $F_1$-PDBP obtains better performed SNNs in TNR only. The iteration number and training time of $F_1$-PDBP is much smaller than that of $F_1$-VPSO.
\begin{table}[ht]
\centering
\caption{KPIs of the SNN$_4$ by PDBP and VPSO on WBC}
\label{tab:KPIs-SNN4-WBC}
\begin{tabular}{l|l|l|l|l|l|l|l}
\hline\small
Algorithms& ACC & $F_1$ & AUC & TPR & TNR & I & T(ms) \\
\hline\hline
 $ERR$-PDBP  & 0.9785  & 0.9694 & 0.9897 & 0.9848 & 0.9753 & 2148 & 14,828\\
 $ERR$-VPSO  & 0.8784  & 0.8184 & 0.9229 & 0.7939 & 0.9229 & 152  & 24,122  \\
 \hline
 $ACC$-PDBP  & 0.9809 &	0.9729 & 0.9876 & 0.9931 & 0.9745 &	334 & 6,057\\
 $ACC$-VPSO  & 0.9766 & 0.9669 & 0.9885 & 0.9903 & 0.9694 & 572 & 23,338 \\
 \hline
 $F_1$-PDBP  & 0.9781 &	0.9687 & 0.9881 & 0.9862 & 0.9738 &	270  & 5,583\\
 $F_1$-VPSO  & 0.9795 & 0.9710 & 0.9901 & 0.9945 & 0.9716 & 1004 & 33,425\\
\hline
\end{tabular}
\end{table}
\paragraph{\bf Performances of the SNN$_7$ obtained by KPI-PDBPs and KPI-VPSOs on SMS\\}
Table \ref{tab:KPIs-SNN7-SMS} shows the performances of SNN$_7$, obtained by KPI-PDBPs and KPI-VPSOs on SMS, the best iteration index (I) when the specific KPI driver or fitness has not changed for $\tau$ (i.e. $20d$), iterations, as well as the training time (T) in the unit of millisecond.

Comparing the performances of SNN$_7$ obtained by ERR-PDBP and ERR-VPSO in Table \ref{tab:KPIs-SNN7-SMS}, ERR-PDBP obtains better performed SNNs in ACC, $F_1$, AUC and TPR, whereas, ERR-VPSO obtains better performed SNNs in TNR only, and the TPR of SNN$_7$, obtained by ERR-VPSO, is too low to be acceptable.  Although the iteration number of ERR-PDBP is much larger than that of ERR-VPSO, the training time of ERR-PDBP is much shorter than that of ERR-VPSO.

Comparing the performances of ACC-PDBP and ACC-VPSO, it can be seen that ACC-PDBP obtains the SNN with slightly better performance in ACC, $F_1$, AUC, and TNR, while ACC-VPSO obtains the SNN with slightly better performance in TPR. The iteration number and training time for ACC-PDBP is smaller than that for ACC-VPSO.

Comparing the performances of SNN$_7$, obtained by $F_1$-PDBP and $F_1$-VPSO, the $F_1$-PDBP obtains the SNN with slightly better performance in ACC, $F_1$, AUC and TNR, whereas $F_1$-VPSO obtains the SNN with slightly better performance in TPR only. The iteration number and training time for $F_1$-PDBP are much smaller than that for $F_1$-VPSO.

All KPI-PDBPs obtain very similar performances, and the order of the training time is: T(ERR-PDBP)>T($F_1$-PDBP)>T(ACC-PDBP).

ERR-VPSO obtains the SNN with lower performances than ACC-VPSO and $F_1$-VPSO, and ACC-VPSO and $F_1$-VPSO obtain the SNNs with similar performances. The order of the training times for KPI-VPSOs is: T(ERR-VPSO) > T($F_1$-VPSO) > T(ACC-VPSO).
\begin{table}[ht]
\centering
\caption{Performance of FNN$_7$ with PDBPs, driven by different KPIs}
\label{tab:KPIs-SNN7-SMS}
\begin{tabular}{l|l|l|l|l|l|l|l}
\hline\small
Algorithms& ACC & $F_1$ & AUC & TPR & TNR & I & T(ms) \\
\hline\hline
ERR-PDBP   & 0.9689	& 0.8775 & 0.9666 & 0.8313 & 0.9902 & 15136 & 637,885\\
ERR-VPSO   & 0.9374   & 0.7137 & 0.9482 & 0.5814 & 0.9925 & 2805  & 4,089,947\\
\hline
ACC-PDBP   & 0.9688 	& 0.8779 & 0.9690 & 0.8358 & 0.9894 & 146   & 89,068\\
ACC-VPSO   & 0.9665   & 0.8703 & 0.9632 & 0.8380 & 0.9864 & 2472  & 2,888,337\\
\hline
$F_1$-PDBP   & 0.9690 	& 0.8776 & 0.9688 & 0.8304 & 0.9904 & 378   &192,380\\
$F_1$-VPSO   & 0.9669   & 0.8718 & 0.9596 & 0.8394 & 0.9867 & 3170  &3,464,413\\
\hline
\end{tabular}
\end{table}
\subsection{Robustness of SNNs with a specific number of hidden neurons}
\subsubsection{Ten-fold Cross Validation of SNNs on WBC}
Now we examine the robustness of the SNN$_4$, obtained by KPI-PDBPs and KPI-VPSOs, using ten-fold cross validation. The robustness of an SNN can be represented by the average performance measure and the standard deviation. A larger average performance measure and a smaller standard deviation represent better robustness. Hence, we use ($a\pm \sigma$ (\%)) to denote an average accuracy ($a$) and a standard deviation ($\sigma$) for the ten runs.
From Table \ref{tab:WBC-tenX}, it can be seen that the average performances of SNN$_4$, obtained by ERR-PDBP, are much better than that of SNNs, obtained by ERR-VPSO and the deviations of SNN$_4$, obtained by ERR-PDBP, are lower than those obtained by ERR-PVSO. Namely, ERR-PDBP produces the SNN with better robustness than ERR-VPSO. The SNN, optimized by ACC-VPSO, has slightly better average performances and smaller deviations than that learned by ACC-PDBP. Namely, the SNN, obtained by ACC-VPSO, has better robustness than that, obtained by ACC-PDBP. The SNN, optimized by $F_1$-VPSO, has slightly better average performances and smaller deviations of performances than that learned by $F_1$-PDBP. Hence, the SNN, produced by $F_1$-VPSO, has better robustness than that obtained by $F_1$-PDBP.
\begin{table}[htp]
\centering
\caption{KPIs and their deviations ($a\pm{\sigma})\%$ of the SNN$_4$ obtained by KPI-PDBPs and KPI-VPSOs}
\label{tab:WBC-tenX}
\begin{tabular}{l|p{35pt}|p{35pt}|p{35pt}|p{35pt}|p{35pt}}
\hline\small
Algorithms& ACC & $F_1$ & AUC & TPR & TNR \\
\hline\hline
ERR-PDBP & 97.73\par$\pm{0.32}$ & 96.75\par$\pm{0.47}$ & 98.82\par$\pm{0.42}$ &	98.38\par$\pm{1.38}$ & 97.38\par$\pm{0.66}$\\
ERR-VPSO & 86.91 \par$\pm{0.78}$ &	79.84\par$\pm{1.17}$ & 92.56\par$\pm{0.97}$ &	75.39\par$\pm{6.35}$ & 92.97\par$\pm{4.06}$\\
\hline
ACC-PDBP & 97.70\par$\pm{0.22}$  &	96.71\par$\pm{0.29}$ & 98.91\par$\pm{0.21}$ &	98.22\par$\pm{1.11}$	& 97.42\par$\pm{0.78}$\\
ACC-VPSO & 97.77\par$\pm{0.14}$ & 96.81\par$\pm{0.19}$ & 99.11\par$\pm{0.13}$ &	98.22\par$\pm{0.55}$ & 	97.53\par$\pm{0.37}$\\
\hline
$F_1$-PDBP &97.54\par$\pm{0.93}$ &	96.53\par$\pm{1.24}$ & 98.77\par$\pm{0.29}$ &	98.87\par$\pm{0.68}$  &	96.83\par$\pm{1.48}$\\
$F_1$-VPSO & 97.98\par$\pm{0.14}$ & 97.13\par$\pm{0.21}$ & 98.94\par$\pm{0.20}$ &	99.17\par$\pm{0.55}$	& 97.36\par$\pm{0.22}$ \\
\hline
\end{tabular}
\end{table}
\subsubsection{Ten-fold Cross Validation of SNNs on SMS}
Now we examine the robustness of the SNN$_7$ obtained  by KPI-PDBPs and KPI-VPSOs, in terms of the results in Table \ref{tab:SMS-tenX}.  It can be seen that, except for TNR, the average performances of SNN$_7$, obtained by ERR-PDBP, are much better than that obtained by ERR-VPSO, and all  deviations of SNN$_7$, obtained by ERR-PDBP is lower than that obtained by ERR-VPSO. TNR of SNN$_7$ obtained by ERR-VPSO is slightly better than that obtained by ERR-PDBP, but the deviation of TNR of SNN$_7$, optimised by ERR-VPSO is larger than that obtained by PDBP. Namely, the SNN$_7$ learned by ERR-PDBP has better robustness than that optimised by ERR-VPSO. ACC-PDBP produces the SNN$_7$ with slightly better average performances and smaller deviations than ACC-VPSO, and only the deviation of AUC of SNN$_7$, learned by ACC-PDBP, is larger than that optimised by ACC-VPSO. Namely, ACC-PDBP produces SNN$_7$ with better robustness than ACC-VPSO. Similarly, except for TPR, other KPIs and their deviations of SNN$_7$, obtained by $F_1$-PDBP, are better than that obtained by $F_1$-VPSO. Hence, $F_1$-PDBP has better robustness than $F_1$-VPSO.
\begin{table}[ht]
\centering
\caption{Performances and deviations ($a\pm{\sigma})\%$ of the SNN$_7$, obtained by KPI-PDBPs and KPI-VPSOs}
\label{tab:SMS-tenX}
\begin{tabular}{l|p{35pt}|p{35pt}|p{35pt}|p{35pt}|p{35pt}}
\hline\small
Algorithms & ACC & $F_1$ & AUC & TPR & TNR\\
\hline\hline
ERR-PDBP &96.85\par$\pm{0.02}$   &87.52\par$\pm{0.16}$ &96.81\par$\pm{0.21}$	&82.50\par$\pm{0.98}$ &99.07\par$\pm{0.14}$\\
ERR-VPSO &93.63\par$\pm{1.0}$ & 70.23\par$\pm{7.10}$ & 94.42\par$\pm{1.06}$ &	57.28\par$\pm{9.74}$ & 99.26\par$\pm{0.49}$ \\
\hline
ACC-PDBP & 96.85\par$\pm{0.04}$ & 87.67\par$\pm{0.10}$  & 96.70\par$\pm{0.34}$ & 83.51\par$\pm{0.64}$ & 98.91\par$\pm{0.13}$\\
ACC-VPSO & 96.56\par$\pm{0.09}$ & 86.59\par$\pm{0.38}$ & 95.84\par$\pm{0.20}$	& 82.88\par$\pm{1.12}$ & 98.68\par$\pm{0.18}$\\	
\hline
$F_1$-PDBP & 96.85\par$\pm{0.04}$ & 87.55\par$\pm{0.20}$ & 96.84\par$\pm{0.26}$ & 82.76\par$\pm{1.06}$ & 99.03\par$\pm{0.17}$ \\
$F_1$-VPSO & 96.61\par$\pm{0.05}$ & 86.87\par$\pm{0.21}$ & 96.12\par$\pm{0.28}$ &	83.68\par$\pm{0.49}$ & 98.61\par$\pm{0.05}$\\
\hline
\end{tabular}
\end{table}
\subsubsection{Impact of different training rates on performances of an SNN for WBC}
\begin{figure}[htp]
\centering
\subfigure[ERR-PDBP and ERR-VPSO]
{\begin{minipage}{0.49\textwidth}
\includegraphics[height=1.8in]{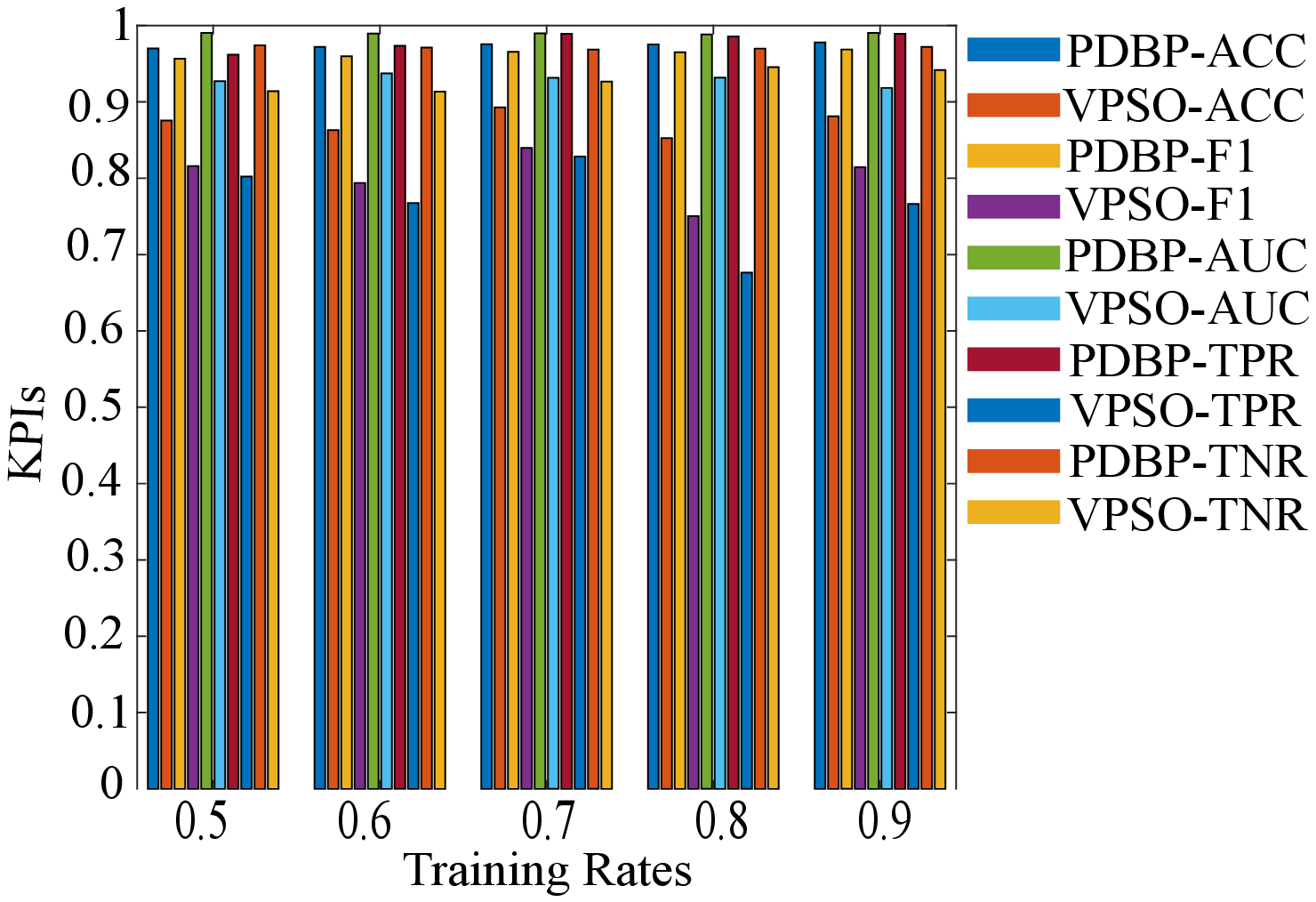}
\end{minipage}}
\subfigure[ACC-PDBP and ACC-VPSO]
{\begin{minipage}{0.49\textwidth}
\includegraphics[height=1.8in]{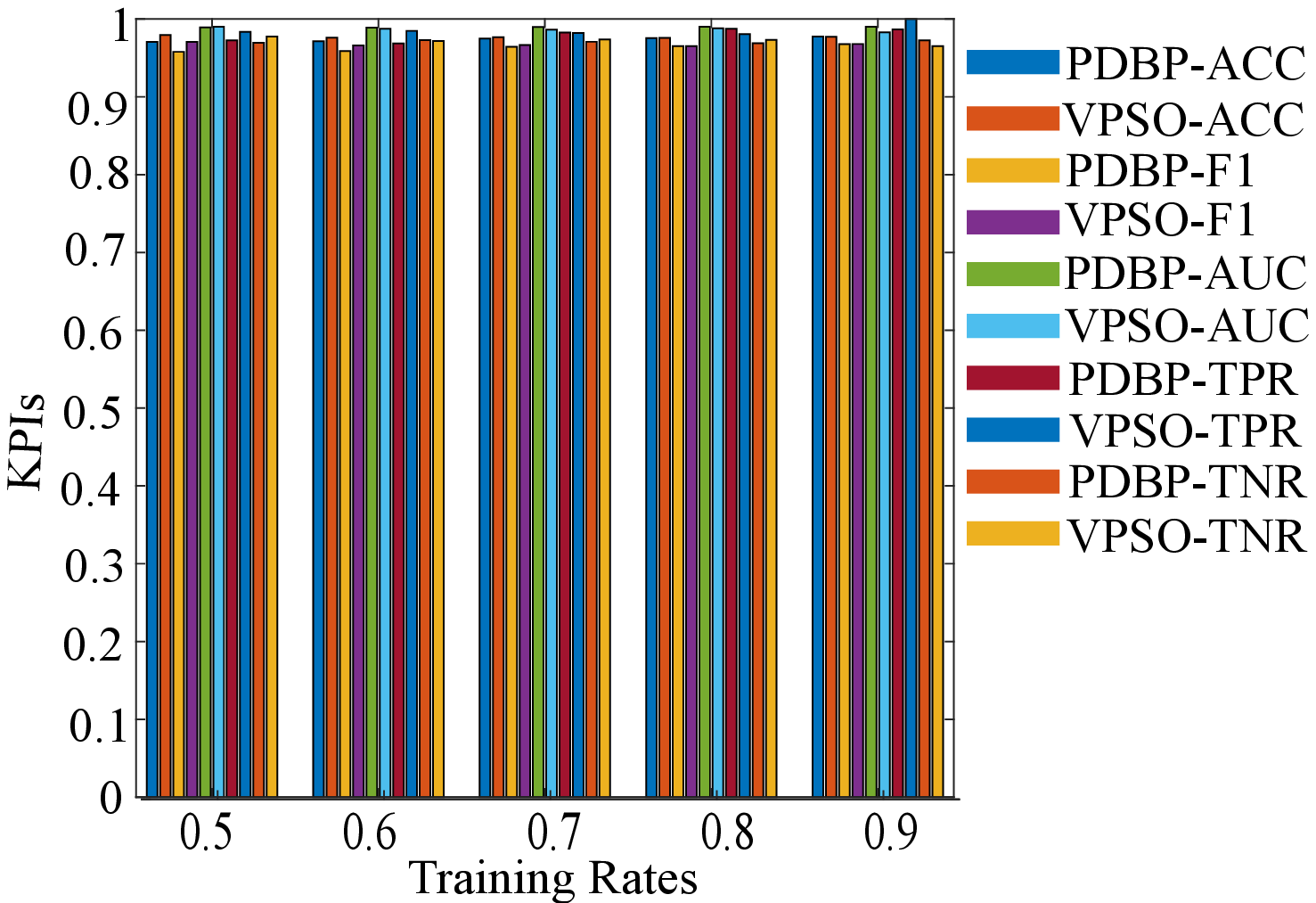}
\end{minipage}}
\subfigure[$F_1$-PDBP and $F_1$-VPSO]
{\begin{minipage}{0.49\textwidth}
\includegraphics[height=1.8in]{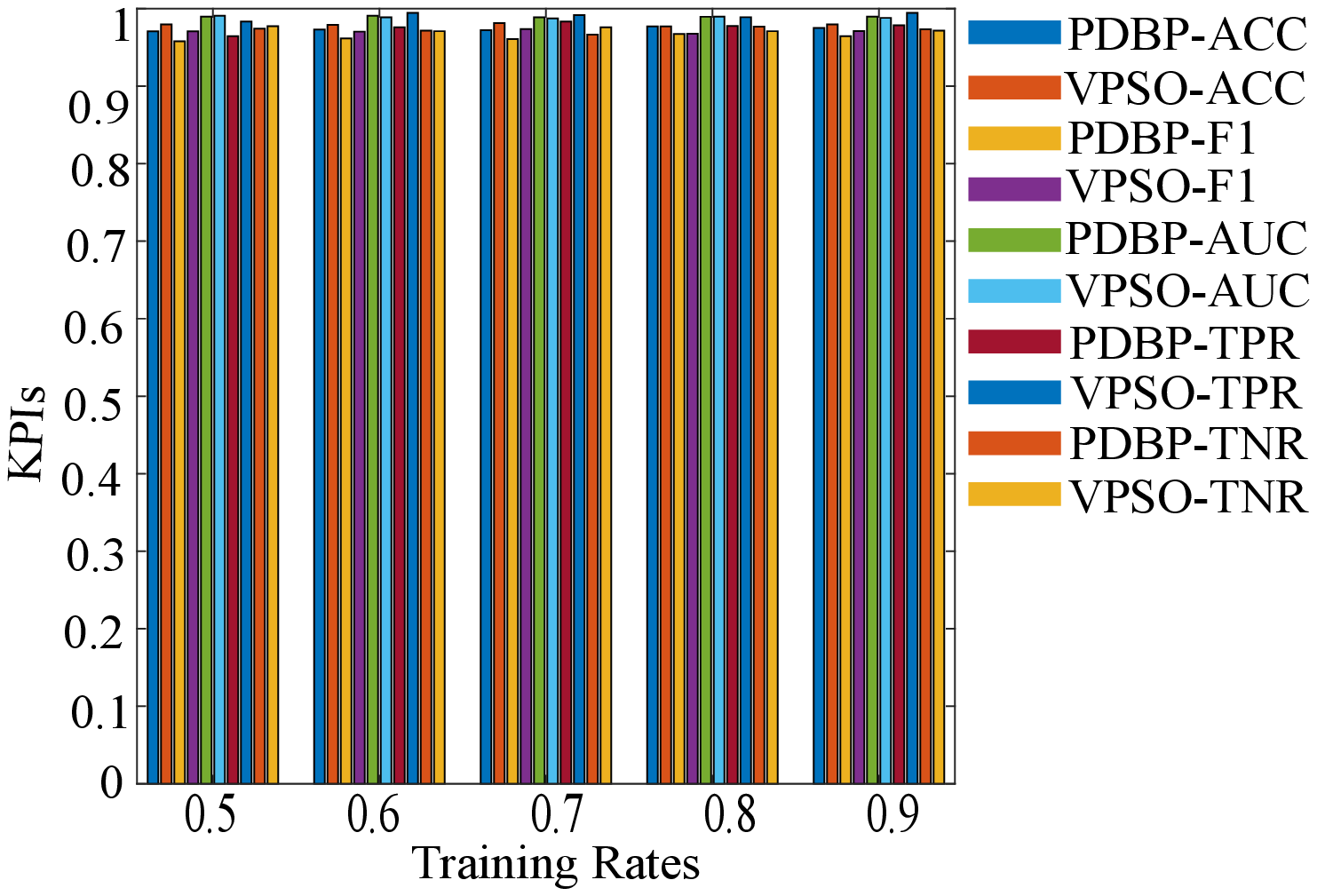}
\end{minipage}}
\caption{KPIs of SNNs, obtained by PDBPs and VPSOs on WBC with different training rates, respectively} \label{fig:TrnRate-WBC}
\end{figure}

To further test the robustness of an SNN, we examine the impact of training rates on the performances of SNN$_5$, trained/optimised by KPI-PDBPs and KPI-VPSOs on WBC with increasing training rates from 0.5 to 0.9.  Figs. \ref{fig:TrnRate-WBC} (a)-(c) show the performances of the SNN$_5$.   From Figs. \ref{fig:TrnRate-WBC} (a)-(c), it can be seen that the performances of SNN$_5$ do not change much as the training rate is changed from 0.5 to 0.9. The results are similar to that in previous experiments for the learnability and robustness (ten fold cross validation). ERR-PDBP always produces the SNN with better performances than ERR-VPSO. ACC-PDBP and ACC-VPSO produces SNNs that have very close performances. Also, $F_1$-PDBP and $F_1$-VPSO obtain SNNs that have very close performances as well.
\subsubsection{Impact of different training rates on KPIs of an SNN for SMS}
\begin{figure}[htp]
\centering
\subfigure[ERR-PDBP and ERR-VPSO on SMS]
{\begin{minipage}{0.49\textwidth}
\includegraphics[height=1.8in]{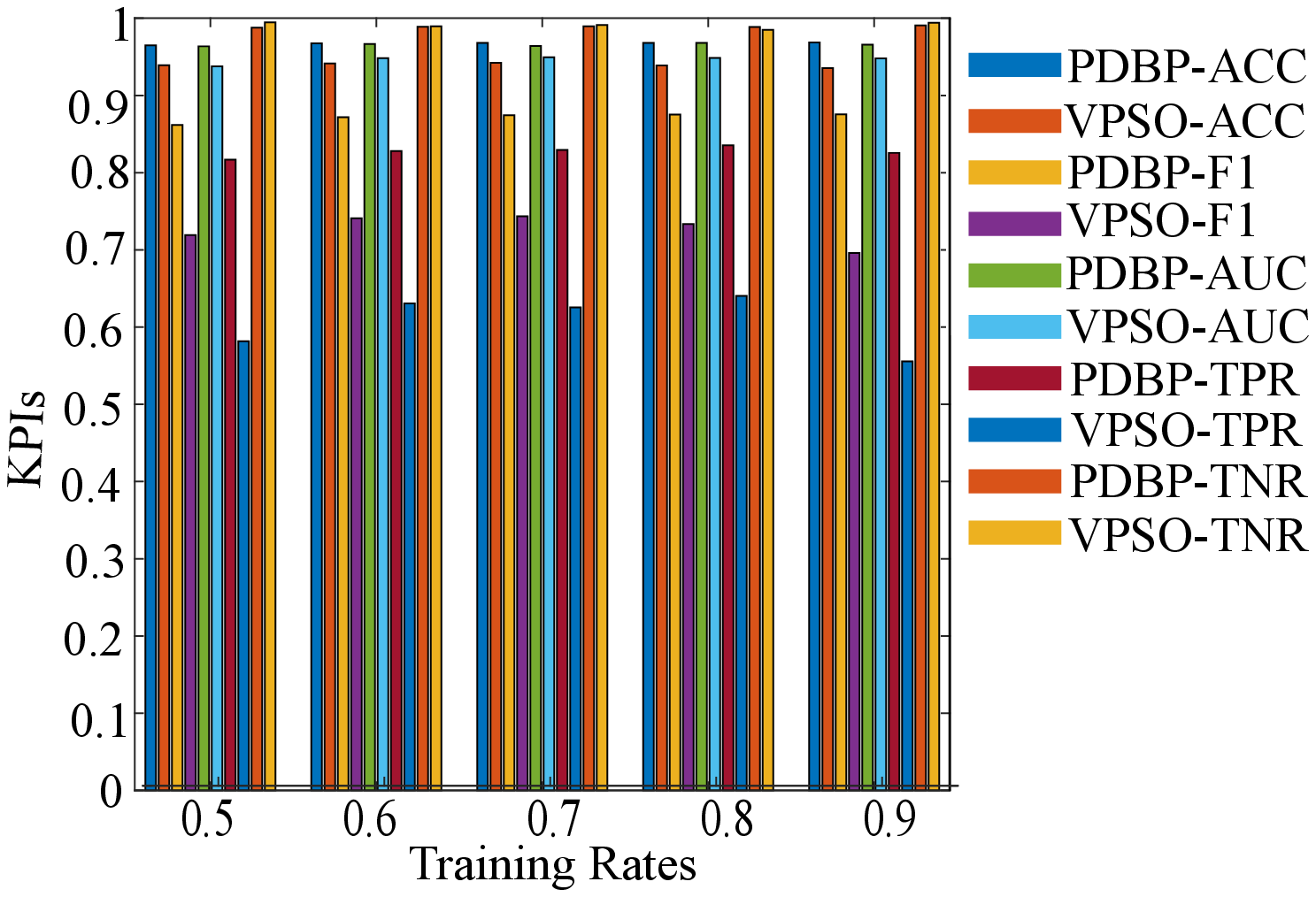}
\end{minipage}}
\subfigure[ACC-PDBP and ACC-VPSO on SMS]
{\begin{minipage}{0.49\textwidth}
\includegraphics[height=1.8in]{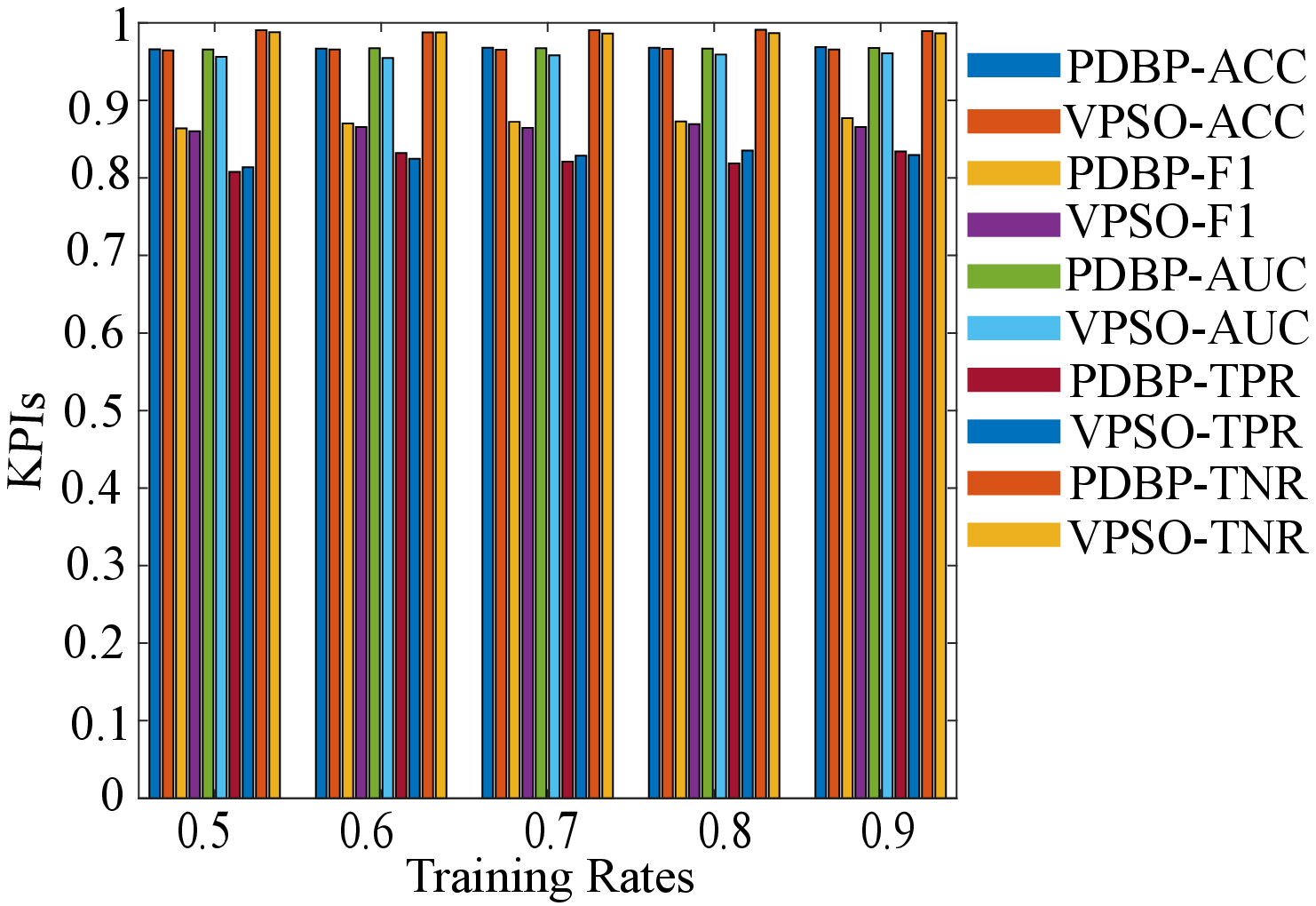}
\end{minipage}}
\subfigure[$F_1$-PDBP and $F_1$-VPSO on SMS]
{\begin{minipage}{0.49\textwidth}
\includegraphics[height=1.8in]{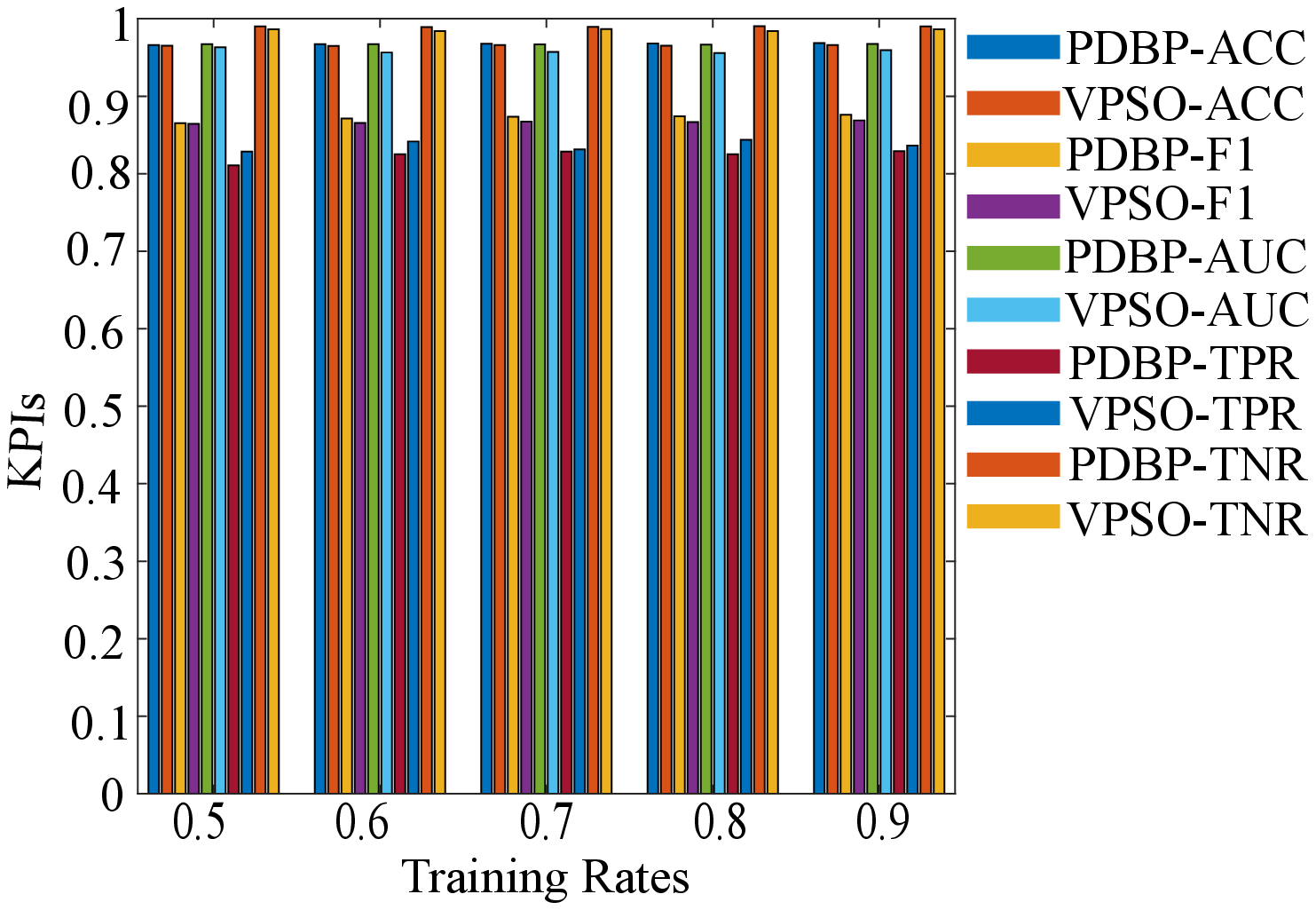}
\end{minipage}}
\caption{KPIs of SNNs, trained by PDBPs and VPSOs on SMS with different training rates, respectively} \label{fig:TrnRate-SMS}
\end{figure}

In the same way, we examine the impact of training rates on the performances of SNN$_8$, obtained by KPI-PDBPs and KPI-VPSOs on SMS with increasing training rates from 0.5 to 0.9.  Figs. \ref{fig:TrnRate-SMS} (a)-(c) show the KPIs of the SNN$_8$. Similar to the results on WBC, the performances of SNN$_8$ do not change much as the training rate changes; ERR-PDBP obtains the SNN with better performances than ERR-VPSO, but they obtain the SNNs with high performance in TNR. ACC-PDBP and ACC-VPSO obtain the SNNs that have similar performances, and performances of learned SNN$_8$ do not change much as the training rate is changed. The performances of SNN$_8$, obtained by KPI-PDBPs and KPI-VPSOs, where KPIs= \{ACC, F$_1$\}, are very close,  and they do not change much as the training rate is changed. However, it can be seen that the $F_1$ and TPR of SNN$_8$, obtained by KPI-PDBPs and KPI-VPSOs, still have a large space to be improved.
\subsubsection{Comparisons of different algorithms on more data sets}
In the same way as in \cite{HL2014}, the experiments were conducted with PDBPs and VPSOs on the 9 data sets for decision making problems from the UCI machine learning repository, including the data set of WBC in the case study. These data sets have been tested with three well-known machine learning approaches, C4.5, Naive Bayes (N.B.) and the Neural Networks (N.N.) in \cite{QZLJ05}, where, WEKA was used to generate the results of J48 (C4.5 in WEKA) unpruned tree, Naive Bayes, and Neural Networks with default parameter settings for ten runs of 50\%-50\% splitting training and test data. These data sets were also assessed by the optimal Linguistic Attribute Hierarchy (LAH), a hierarchy of Linguistic Decision Trees \cite{HL2014}. The SNNs, obtained by ACC-PDBP and ACC-VPSO, are assessed on these data sets for comparison.

Table \ref{tab:comparison} lists the dimension number ($n$) and the sample number ($N$) of the tested databases. The accuracies and standard deviations for different models are presented in the form of ($a\pm \sigma$ (\%)), where $a$ is an average accuracy, and $\sigma$ is a standard deviation.

The running times of the evolution processes are very long for very high-dimensional databases. Therefore, in the same way in \cite{HL2014}, for those data sets that have very high-dimensional input spaces (e.g., the data sets of Ionosphere and Sonar), a pre-processing was done. All attributes are sorted on a non-increasing order of information gains relative to the goal classes, and the first 12 attributes of the data are selected to construct an SNN to map the relationship between input attributes and the goal in the reduced dimensions of data. The number of hidden neurons is set to half of the feature number after pre-processing.

From Table \ref{tab:comparison}, it can be seen that the SNNs, obtained by ACC-PDBP, win for almost all data sets, except for the data set of Hepatitis, the optimal LAH is the best, and ACC-PDBP is at the second place.  The deviation of accuracy of SNN$_4$, obtained by ACC-PDBP is ranked at the 1st or the 2nd place, except for Ionosphere, the deviation of accuracies for SNN$_4$, obtained by ACC-PDBP, is at the third place among all models.
\begin{table}[ht!]
\centering
\caption{Average ordinary accuracies (\%) and standard
deviations obtained by 3 well-known machine learning approaches,
LAHs, ACC-PDBP and ACC-VPSO} \label{tab:comparison} \scriptsize
\begin{tabular}{p{30pt}|p{15pt}|p{15pt}|p{35pt}|p{35pt}|p{35pt}|p{35pt}|p{35pt}|p{35pt}}
\hline
$Data$ &$n$ & $N$ & C4.5 &N.B.&N.N.&LAH&ACC-PDBP&ACC-VPSO\\
\hline
BreastC.      & 9 & 286 &69.16\par$\pm{4.14}$ &71.26\par$\pm{2.96}$ &66.50\par$\pm{3.48}$  &71.77\par$\pm{2.06}$ &{\bf 78.67}\par$\pm{1.79}$ &75.59\par$\pm{1.57}$\\
WBC           & 9 & 699&94.38\par$\pm{1.42}$ &96.28\par$\pm{0.73}$ &94.96\par$\pm{0.80}$  &96.67\par$\pm0.20$ &{\bf 97.14}\par$\pm{0.34}$ &95.64\par$\pm{0.49}$\\
Heart-c       & 13& 303&75.50\par$\pm{3.79}$ &84.24\par$\pm{2.09}$ &79.93\par$\pm{3.99}$  &82.81\par$\pm{4.25}$ &{\bf 87.03}\par$\pm{1.13}$ &81.35\par$\pm{2.35}$\\
Heart-s.      & 13& 270&75.78\par$\pm{3.16}$ &84.00\par$\pm{1.68}$ &78.89\par$\pm{3.05}$  &84.85\par$\pm{2.31}$ &{\bf 88.37}\par$\pm{1.21}$ &81.35\par$\pm{2.35}$\\
Hepatitis     & 19& 155&76.75\par$\pm{4.68}$ &83.25\par$\pm{3.99}$ &81.69\par$\pm{2.48}$  &{\bf 94.84}\par$\pm{1.01}$ &85.23\par$\pm{2.36}$ &81.23\par$\pm{3.23}$\\
Ionosphere    & 34& 351&89.60\par$\pm{2.13}$ &82.97\par$\pm{2.51}$ &87.77\par$\pm{2.88}$  &89.80\par$\pm{1.63}$ &{\bf 93.90}\par$\pm{1.65}$  &91.88\par$\pm{0.75}$\\
Liver         & 6 & 345&65.23\par$\pm{3.86}$ &55.41\par$\pm{5.39}$ &66.74\par$\pm{4.89}$  &58.46\par$\pm{0.76}$ &{\bf 73.62}\par$\pm{1.97}$ &72.35\par$\pm{2.06}$\\
Diabetes      & 8 & 768&72.16\par$\pm{2.80}$ &75.05\par$\pm{2.37}$ &74.64\par$\pm{1.41}$  &76.07\par$\pm{1.33}$ &{\bf 79.47}\par$\pm{1.12}$ &70.66\par$\pm{1.19}$\\
Sonar         & 60& 208&70.48\par$\pm{0.00}$ &70.19\par$\pm{0.00}$ &81.05\par$\pm{0.00}$  &74.81\par$\pm{4.81}$ &{\bf 88.75}\par$\pm{1.68}$ &69.33\par$\pm{2.26}$\\
\hline
\end{tabular}
\end{table}
\section{Conclusions}
To save computing resources, a shallow neural network (SNN) may be useful in adaptive edge intelligence, and the number of hidden neurons of an SNN could be adaptive to the attribute number in the problem space. Hence, it is important to examine the learnability and robustness of SNNs. The contributions of the research are summarised as follows:
(1) A performance driven BP (PDBP) algorithm to reduce the training time and mitigate overfitting, and a variant of PSO (VPSO) to optimise the weights of an SNN, are developed.
(2) The learnability of SNNs are validated through a tight force heuristic algorithm (PDBP) and a loose force meta-heuristic algorithm (VPSO).
(3) An incremental approach is used to examine the impact of hidden neuron numbers on the performances of an SNN, trained by KPI-PDBPs, where KPIs=\{ERR, ACC and $F_1$\}. The number of hidden neurons of an SNN depends on the non-linearity of the training data. The experimental results show that the performance of the SNN may not be further improved very much, if the number of hidden neurons is over half of the feature number in the problem space. It should be noticed that a local optimum needs to be avoided for both PDBPs and VPSOs.
(4) The robustness of SNNs are examined through changing the training rates. The performances of SNNs, obtained by KPI-PDBPs and KPI-VPSOs, do not change much for different training rates. Comparing with other classic machine learning algorithms, such as C4.5, NB and NN in literature, the SNNs obtained by ACC-PDBP win for almost all tested data sets.
Table \ref{tab:summary} summarises the learnability and the robustness of SNNs, obtained by KPI-PDBPs and KPI-VPSOs on the two benchmark data sets.
\begin{table}[ht]
\centering
\caption{The learnability and robustness of SNNs, obtained by PDBPs and VPSOs}
\label{tab:summary}
\begin{tabular}{l|l|l|l|l|l|l|l|l}
\toprule
& \multicolumn{4}{c|}{Learnability}&\multicolumn{4}{c}{Robustness} \\
\cline{2-9}
Algorithms & \multicolumn{2}{c|}{WBC} & \multicolumn{2}{c|}{SMS}  & \multicolumn{2}{c|}{WBC} & \multicolumn{2}{c}{SMS}\\
\cline{2-9}
              & KPIs & T & KPIs &T  &  KPIs & $\sigma$  & KPIs & $\sigma$  \\
\midrule[1pt]
ERR-PDBP & \checkmark  & \checkmark & \checkmark & \checkmark & \checkmark  & \checkmark & \checkmark & \checkmark \\
\midrule[0.3pt]
ERR-VPSO &             &            &            &            &             &            &            &            \\
\midrule[1pt]
ACC-PDBP &  \checkmark &\checkmark  &\checkmark  &\checkmark  &             &            &\checkmark  &\checkmark \\
\midrule[0.3pt]
ACC-VPSO &             &            &            &            &\checkmark   &\checkmark  &            &  \\  	
\midrule[1pt]
$F_1$-PDBP  &          & \checkmark &\checkmark  & \checkmark &             &            &\checkmark  &\checkmark  \\
\midrule[0.3pt]
$F_1$-VPSO  &\checkmark&            &            &            & \checkmark  &\checkmark  &            &            \\
\bottomrule
\end{tabular}
\end{table}

Finally, the research results show that a shallow neural network may align with the property of the human brain: when an individual gets the first impression to a thing, without new information, the individual cannot change the impression,  and a tight force search, which takes the nature of the problem, may be easier to close to the truth than a loose force search.

\section*{Acknowledgements}
This research is sponsored by National Natural Science Foundation of China (61903002) and the key project of Natural Science in Universities in Anhui Province, China (KJ2018A0111).

\end{document}